\documentclass[sn-mathphys]{sn-jnl}

\usepackage{graphicx}%
\usepackage{multirow}%
\usepackage{amsmath,amssymb,amsfonts}%
\usepackage{amsthm}%
\usepackage{mathrsfs}%
\usepackage[title]{appendix}%
\usepackage{xcolor}%
\usepackage{textcomp}%
\usepackage{manyfoot}%
\usepackage{booktabs}%
\usepackage{algpseudocode}%
\usepackage{listings}%

\theoremstyle{thmstyleone}%
\theoremstyle{thmstyletwo}%

\theoremstyle{thmstylethree}%

\usepackage{amssymb}
\usepackage{xspace}
\usepackage{amsmath}
\usepackage{multirow}
\usepackage{array}
\usepackage{booktabs}
\usepackage{cleveref}
\usepackage{balance}
\usepackage[linesnumbered, ruled, vlined]{algorithm2e}
\usepackage{subfigure}

\newcommand{\ourmethod}{\textbf{C}\textsc{Sincere}\xspace}

\crefname{equation}{Eq.}{Eqs.}
\crefname{table}{Table}{Tables}
\crefname{figure}{Fig.}{Figures}
\crefname{algorithm}{Algorithm}{Algorithms}
\crefname{algocf}{Algorithm}{Algorithms}
\crefname{section}{Sec.}{Secs.}

 \usepackage{amsthm}

\begin{document}

\title[Article Title]{Contrastive Sequential Interaction Network Learning on Co-Evolving Riemannian Spaces}


\author*[1]{\fnm{Li} \sur{Sun}}\email{ccesunli@ncepu.edu.cn}

\author[2]{\fnm{Junda} \sur{Ye}}\email{jundaye@bupt.edu.cn}

\author[3]{\fnm{Jiawei} \sur{Zhang}}\email{jiwzhang@ucdavis.edu}

\author[4]{\fnm{Yong} \sur{Yang}}\email{ yangyongxp@163.com}

\author*[5]{\fnm{Mingsheng} \sur{Liu}}\email{liums601001@sina.com}

\author[2]{\fnm{Feiyang} \sur{Wang}}\email{fywang@bupt.edu.cn}

\author[6]{\fnm{Philip S.} \sur{Yu}}\email{psyu@uic.edu}

\affil*[1]{\orgname{North China Electric Power University}, \orgaddress{\postcode{102206}, \state{Beijing}, \country{China}}}

\affil[2]{\orgname{Beijing University of Posts and Telecommunications}, \orgaddress{\postcode{100876}, \state{Beijing}, \country{China}}}

\affil[3]{\orgname{University of California, Davis}, \orgaddress{\postcode{95616}, \state{CA}, \country{USA}}}

\affil[4]{\orgname{State Grid Handan Electric Power Supply Company}, \orgaddress{\postcode{056000}, \city{Handan}, \state{Hebei}, \country{China}}}

\affil[5]{\orgname{Shijiazhuang Institute of Railway Technology}, \orgaddress{\postcode{050041}, \city{Shijiazhuang}, \city{Hebei}, \country{China}}}

\affil[6]{\orgname{University of Illinois at Chicago}, \orgaddress{\postcode{60607}, \state{IL}, \country{USA}}}



\abstract{
The sequential interaction network usually find itself  in a variety of applications, e.g., recommender system. Herein, inferring future interaction is of fundamental importance, and previous efforts are mainly focused on the dynamics  in the classic zero-curvature Euclidean space. 
Despite the promising results achieved by previous methods, a range of significant issues still largely remains open: 
On the bipartite nature, is it appropriate to place user and item nodes in one identical space regardless of their inherent difference? 
On the network dynamics, instead of a fixed curvature space, will the representation spaces evolve when new interactions arrive continuously? 
On the learning paradigm, can we get rid of the label information costly to acquire? 
To address the aforementioned issues, we propose a novel \textbf{\underline{C}}ontrastive model for \textbf{\underline{S}}equential \textbf{\underline{I}}nteraction \textbf{\underline{N}}etwork learning on \textbf{\underline{C}}o-\textbf{\underline{E}}volving \textbf{\underline{R}}i\textbf{\underline{E}}mannian spaces, \ourmethod. 
To the best of our knowledge, we are the first to introduce a couple of co-evolving representation spaces, rather than a single or static space, and propose a co-contrastive learning for the sequential interaction network.
In \ourmethod, we formulate a Cross-Space Aggregation for message-passing across representation spaces of different Riemannian geometries,
and design a Neural  Curvature Estimator based on Ricci curvatures for modeling the space evolvement over time.
Thereafter, we present a Reweighed Co-Contrast between the temporal views of the sequential network,
so that the couple of Riemannian spaces interact with each other for the interaction prediction without labels. 
Empirical results on 5 public datasets show the superiority of \ourmethod over the state-of-the-art methods. 
}




\maketitle

\section{Introduction}
Interaction prediction for sequential interaction networks (represented as temporal bipartite graphs) is essential in a wide spectrum of applications, e.g.,  ``Guess You Like'' in recommender systems \citep{he2014predicting,2021LIME}, 
``Related Searches'' in search engines \citep{pengTOIS} 
and ``Suggested Posts'' in social networks \citep{HTGN, CAW}. 
Specifically, in the e-commerce platforms, the trading or rating behaviors indicate the interactions between users (i.e., purchasers) and items (i.e., commodities), and thus predicting interactions helps improve the quality and experience of  recommender system. 
In a social media, the cases that a user clicks on or comments on the posts correspond to user-item interactions, 
and blocking interactions from malicious posts (such as the promotion of drugs) is significant for social good especially for the care of teenagers \citep{pengPAMI}.

Graph representation learning, which represents nodes as low-dimensional embeddings, supports and facilitates interaction prediction. 
\textbf{Which space is appropriate to accommodate the embeddings} is indeed a fundamental question. 
To date, the answer of most previous works is the (zero-curvature) Euclidean space. 
Nevertheless, the recent advances show that Euclidean space is usually not a good answer, especially for the graphs presenting dominant hierarchical/scale-free structures \citep{chami2019hyperbolic}. 
The Riemannian space has emerged as an exciting alternative.
For instance, Riemannian spaces with negative curvatures \footnote{In Riemannian geometry, the negative curvature space is termed as the hyperbolic space, and positive curvature space is termed as the spherical space.} are well aligned with hierarchical structures while the positive curvature ones for cyclical structures \citep{mathieu2019continuous, HAN, ZhangWSLS21}. 
In fact, Euclidean space is a special case of Riemannian space with zero curvature.
In the context of graph representation learning, the hyperbolic space was first introduced in  \citet{nickel2017poincare,  HNN}, while \citet{defferrard2020deepsphere, rezende2020normalizing} explore the representation learning in spherical spaces. 
More specifically, in the representation learning on sequential interaction networks, most of the previous studies model the sequence of user-item interactions and learn the embeddings in Euclidean space \citep{tgsrec, BiGI, deepcoevolve, CTDNE, deepred, latentcross}.
The previous studies ignore the complex underlying structures in sequential interaction networks, and thus motivate us to \emph{study sequential interaction network learning in Riemannian space with more generic expressive capacity}.

Herein, we summarize the major shortcomings of previous sequential interaction network learning methods as follows:
\begin{itemize}
 \item  The first issue is on the \textbf{bipartite nature}.
To the best of our knowledge, all  existing studies in the literature simply set \textit{two different} types of nodes (users and items) in \textit{one identical} space.
It is counter-intuitive and ambiguous. 
For instance, viewers and films are two kinds of nodes with totally different characters and distributions \citep{sreejith2016forman, bachmann20a}. 
Also, we give another motivated example by empirically investigating on MOOC and Wikipedia. The results are shown in \cref{study case figure} and \cref{study case table}, where $\delta$ quantifies the shape of the structure.  
Obviously, we find that the users and items are different from each other in terms of both $\delta$-hyperbolicity and degree distribution.
Thus, rather than a single space, it is more rational to model the users and items in two different spaces.
Riemannian geometry provides the notion of curvature to distinguish the structural pattern between different spaces.
Unfortunately, the representation learning over two different spaces (e.g., how to pass the message cross different spaces) largely remains open.



\begin{table}[ht]
    \caption{The average $\delta$-hyperbolicity \citep{Gromov1987} of user and item subgraphs on MOOC and Wikipedia dataset along the timeline.}
    \begin{tabular}{l | c c c |l | c c c}
        \toprule
        \textbf{Dataset} & \textbf{Timeline} & \textbf{User} & \textbf{Item} & \textbf{Dataset} & \textbf{Timeline} & \textbf{User} & \textbf{Item}\\
        \hline
        \multirow{3}{*}{MOOC} & start & 0.77 & 0.50    & \multirow{3}{*}{Wikipedia} & start & 1.04 & 1.16 \\
         & middle & 1.0 & 0.67     &   & middle & 1.10 & 1.30 \\ 
         & end & 1.0 & 0.13        &     & end & 1.33 & 1.40 \\   
        \bottomrule
    \end{tabular}
        \label{study case table}
\end{table}

\begin{figure}[ht]
 \vspace{-0.1in}
    \centering
    \includegraphics[width=0.7\linewidth]{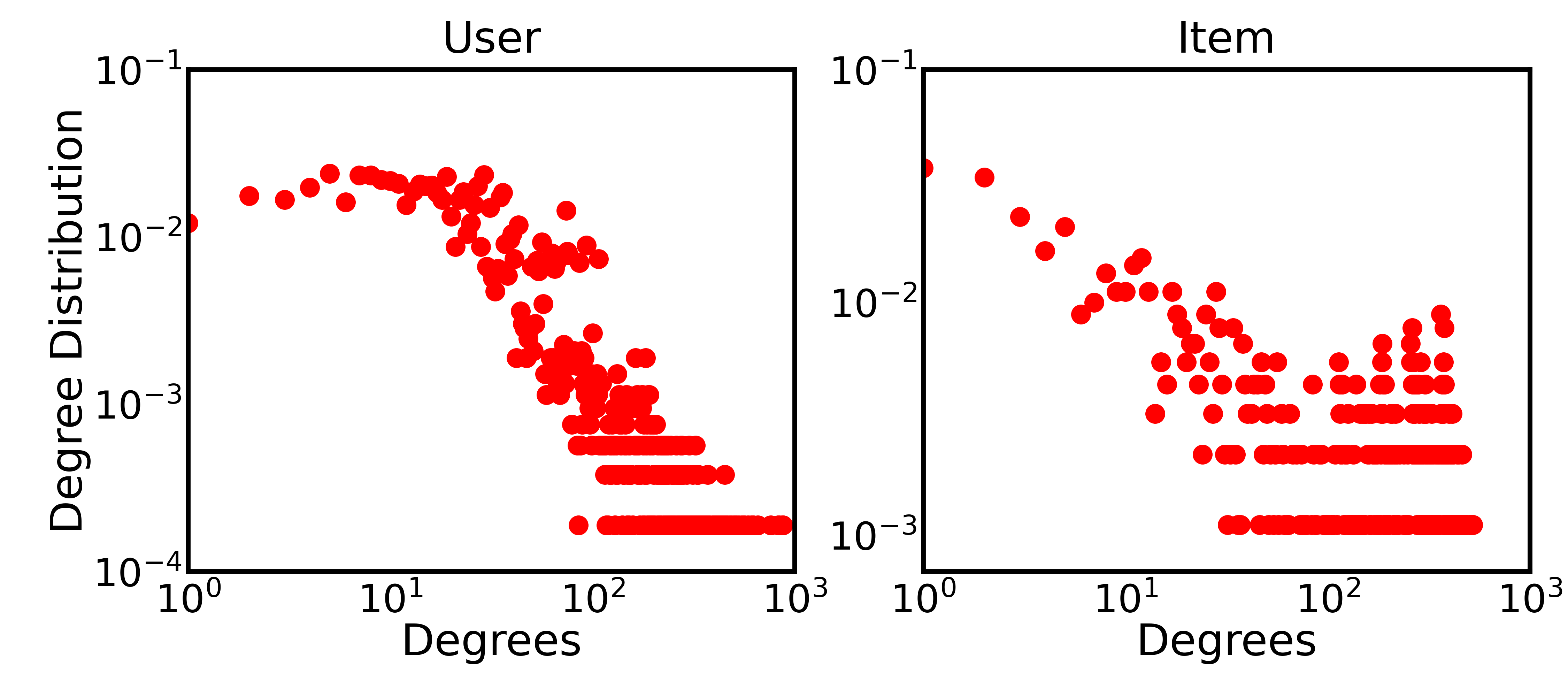}
        \vspace{-0.1in}
    \caption{User and item degree distribution on Wikipedia at the end interval.}
    \label{study case figure}
\end{figure}

\item The second issue is on the \textbf{network dynamics.} We notice that \citet{hyperml,HTGN} represent sequential interaction networks on the hyperbolic manifolds \footnote{We use manifold and space interchangeable throughout this paper.} very recently. 
They still assume the space is static same as the previous studies.
 The fact that the interaction network constantly \textit{evolves} over time is largely ignored.
It is also evidenced in the example in \cref{study case figure} and \cref{study case table}.  Both $\delta$-hyperbolicity and degree distribution vary over time.
 In the language of  Riemannian geometry, the previous studies attempt to learn user/item embeddings on a  fixed curvature space, either Euclidean space or hyperbolic ones. 
Rather than a fixed curvature space, it calls for an evolving curvature space to manifest the inherent dynamics  of the sequential interaction network, where the new interactions continuously arrive. 
The challenge lies in how to estimate  curvature to model the space evolvement as it still lacks effective estimator in the literature. 
\end{itemize}
In this paper, we argue that the representation space for sequential interaction network needs to model the difference between users and item (the first issue) and the evlovement over time (the second issue).
\begin{itemize}
\item The third issue is on the \textbf{learning paradigm.} The graph models is typically trained with abundant labels. 
Undoubtedly, the labels are expensive to acquire, and the reliability of the labels is sometimes questionable, especially for the case that the interactions continuously arrive. 
In the literature, the self-supervised learning on graphs without labels is roughly divided into generative methods and contrastive methods.
The generative methods require a carefully designed decoder for data reconstruction.
On the contrary, contrastive methods are free of decoder, and acquire knowledge by distinguishing the positive pairs from the negative pairs. 
Recently, \emph{contrastive learning} has achieved the state-of-the-art performance for the typical graphs (e.g., social networks and citation networks), but it \emph{still remains open for sequential interaction networks}.
\end{itemize}
Besides, most of the previous work \citep{jodie, hili, time-lstm, tlstm} consider that the interactions would only explicitly link the nodes of different type in the bipartite interaction graph 
while ignoring the implicit interaction among the same type of nodes. 
Indeed, it calls for new method to model such explicit and implicit impacts among the nodes.


To address the issues above, we propose a  \textbf{\underline{C}}ontrastive  \textbf{\underline{S}}equential \textbf{\underline{I}}nteraction \textbf{\underline{N}}etwork learning model on \textbf{\underline{C}}o-\textbf{\underline{E}}volving \textbf{\underline{R}}i\textbf{\underline{E}}mannian manifolds (\ourmethod). 
We first present a Co-evolving GNN to address the first and second issues.
For the first issue, we propose to model the users and items in two different $\kappa$-stereographic spaces, i.e.,  Riemannian user space and Riemannian item space. 
The user space and item space are linked with the Euclidean tangent space, modeling the temporal interactions.
Co-evolving GNN utilizes \emph{Cross-Space Aggregation} for the message passing across different Riemannian spaces.
For the second issue, we learn the  temporal evolvement of user/item space curvatures via a neural curvature estimator (\emph{CurvNN}). 
Co-evolving GNN utilizes \emph{CurvNN} for both  user and item spaces, so that they co-evolve with each other over time.
Rather a single or static representation space, we for the first time embed the sequential interaction network into co-evolving Riemannian manifolds.

Our preliminary work above \citep{YeJD23WWW} has addressed the first and second issues, and this paper
 has equipped the original learning model on co-evolving Riemannian manifolds in \citep{YeJD23WWW} with a novel contrastive learning approach.
Specifically, this full version further consider the issue of self-supervised learning for the sequential interaction network (i.e., the third issue).
To this end, we introduce a Riemannian co-Contrastive learning for sequential interaction networks, where we propose to contrast between the temporal views generated in the evolvement. 
In the co-Contrastive learning, the novelty lies in that the users are co-contrasted with both users and items, and vice versa, so that Riemannian user space and Riemannian item space interact with each other. 
In the meanwhile, we pay more attention to both hard positive samples and hard negative samples with the reweighing mechanism.
Finally, interacting between user space and item space, \ourmethod predicts future interactions without label information.

Overall, the noteworthy contributions of our work are summarized as follows:
\begin{itemize}
    \item \emph{Problem}. 
    We rethink the bipartite nature, network dynamics and learning paradigm of sequential interaction network learning.
    It is the first attempt to introduce the co-evolving Riemannian manifolds, to the best of our knowledge.
    \item \emph{Model}. 
    We propose a novel co-evolving GNN with the cross-space aggregation, which represents the users and items in two different $\kappa$-stereographic spaces, co-evolving over time with the parameterized curvatures.
    \item \emph{Learning Paradigm}. 
    We propose a novel contrastive learning approach for sequential interaction networks, which interplays the co-evolving user and item space for interaction prediction.
    \item \emph{Experiment}.  
    Empirical results  on 5 public datasets show the superiority of the proposed approach against the strong baselines.
\end{itemize}

\noindent \textbf{Roadmap.} The rest parts are organized as follows:
We introduce the preliminary mathematics and formulate the studied problem in Sec. 2. 
To address this problem, we present the co-evolving graph neural network in Sec. 3, and the reweighted co-contrastive learning in Sec. 4.
The empirical results of the proposed approach are reported in Sec. 5.
We summarize the related work in Sec. 6, and finally conclude our work in Sec. 7.

\section{Preliminaries \& Problem Formulation}

In this section, we first introduce the preliminary mathematics on Riemannian manifold and the notion of curvature. Then, we formulate the studied problem of \emph{self-supervised Riemannian sequential interaction network learning}.

\subsection{Riemannian Manifold}
A smooth manifold $\mathcal{M}$ is said to be a Riemannian manifold if it is endowed with a Riemannian metric $g$. The Riemannian metric is characterized as the positive-definite inner product defined on $\mathcal{M}$'s tangent space $g_{\boldsymbol{x}}: \mathcal{T}_{\boldsymbol{x}}\mathcal{M} \times \mathcal{T}_{\boldsymbol{x}}\mathcal{M} \rightarrow \mathbb{R}$. 
Concretely, the tangent space $\mathcal{T}_{\boldsymbol{x}}\mathcal{M}$ is associated with a point $\boldsymbol{x}$ on the manifold, approximating the locality of the geometry with Euclidean space.
That is, a Riemannian manifold is a tuple $(\mathcal{M}, g)$ where $g_{\boldsymbol{x}}(\boldsymbol{v}, \boldsymbol{v}) \geq 0, g_{\boldsymbol{x}}(\boldsymbol{u}, \boldsymbol{v}) = g_{\boldsymbol{x}}(\boldsymbol{v}, \boldsymbol{u})$ with $\boldsymbol{x} \in \mathcal{M}$ and $ \boldsymbol{u}, \boldsymbol{v} \in \mathcal{T}_{\boldsymbol{x}}\mathcal{M}$.


\subsection{The $\kappa-$stereographic Model}

In Riemannian geometry, there exists three kinds of isotropic spaces: hyperbolic, spherical and Euclidean space. 
Note that, the vectors in either the hyperbolic or the spherical space cannot be operated as the way we are familiar with in Euclidean space.
The $\kappa$-stereographic model \citep{bachmann20a}  unifies the vector operations of the aforementioned three kinds of isotropic spaces with gyrovector formalism \citep{ungar2008gyrovector}.  (Notations: Bold lowercase $\boldsymbol{x}$ and bold uppercase $\boldsymbol{X}$ denote the vector and matrix, respectively.)

Without loss of generality, we consider the $n$-dimensional  model ($n \ge 1$). 
The \textbf{$\kappa$-stereographic model} is a smooth manifold $\mathcal{M}_{\kappa}^{n}=\left\{\boldsymbol{x} \in \mathbb{R}^{n} \mid-\kappa\|\boldsymbol{x}\|_{2}^{2}<1\right\}$ equipped with a Riemannian metric $g_{\boldsymbol{x}}^{\kappa}=\left(\lambda_{\boldsymbol{x}}^{\kappa}\right)^{2} \boldsymbol{I}$, where $\kappa \in \mathbb{R}$ is the (sectional) curvature and $\lambda_{\boldsymbol{x}}^{\kappa}=2\left(1+\kappa\|\boldsymbol{x}\|_2^{2}\right)^{-1}$ is the conformal factor defined on the point $\boldsymbol{x}$. 
More specifically, when $\kappa > 0$, $\kappa$-stereographic model shift to the spherical space (i.e., the stereographic projection of the hypersphere model. When $\kappa < 0$,  $\kappa$-stereographic model shift to the hyperbolic space (i.e., the Poincar{\'{e}} ball model with the radius of $1/\sqrt{-\kappa}$), and $\kappa$-stereographic model becomes Euclidean with $\kappa =0$ as a special case.
Now, we introduce the gyrovector  operations as follows.

\noindent \textbf{M\"{o}bius Addition.} In the gyrovector  formalism, M\"{o}bius addition $\oplus_{\kappa}$ of two points $\boldsymbol{x}, \boldsymbol{y} \in \mathcal{M}_{\kappa}^{n}$ is defined as follows:
\vspace{-0.1in}
\begin{equation} \label{mobius addition}
    \begin{aligned}
        \boldsymbol{x} \oplus_{\kappa} \boldsymbol{y} &=\frac{\left(1-2 \kappa\langle\boldsymbol{x}, \boldsymbol{y}\rangle-\kappa\|\boldsymbol{y}\|_{2}^{2}\right) \boldsymbol{x}+\left(1+\kappa\|\boldsymbol{x}\|_{2}^{2}\right) \boldsymbol{y}}{1-2 \kappa\langle\boldsymbol{x}, \boldsymbol{y}\rangle+\kappa^{2}\|\boldsymbol{x}\|_{2}^{2}\|\boldsymbol{y}\|_{2}^{2}}.
    \end{aligned}
\end{equation}
Note that, the M\"{o}bius addition is  non-associative.

\noindent  \textbf{M\"{o}bius Scaling and Matrix-Vector Multiplication.} 
Scaling a $\kappa$-stereographic vector $\boldsymbol{x} \in \mathcal{M}_{\kappa}^{n} \backslash \left\{\boldsymbol{o}\right\}$ is defined with $\otimes_{\kappa}$ in Eq. (\ref{mobius scalar multiplication}), where $r \in \mathbb{R}$ is the scaling factor. The matrix-vector multiplication  of any $\boldsymbol{M} \in \mathbb{R}^{m \times n}$ is given in Eq. (\ref{mobius matrix multiplication}) as follows:
\vspace{-0.1in}
\begin{align}
    r \otimes_{\kappa} \boldsymbol{x} = &\frac{1}{\sqrt{\kappa}} \tan_\kappa \left(r \tan_\kappa ^{-1}(\sqrt{\kappa}\|\boldsymbol{x}\|_{2})\right) \frac{\boldsymbol{x}}{\|\boldsymbol{x}\|_{2}}, \label{mobius scalar multiplication} \\
    \boldsymbol{M} \otimes_{\kappa} \boldsymbol{x} =(1 / \sqrt{\kappa}) &\tan_\kappa \left(\frac{\|\boldsymbol{M} \boldsymbol{x}\|_2}{\|\boldsymbol{x}\|_2} \tan_\kappa ^{-1}(\sqrt{\kappa}\|\boldsymbol{x}\|_2)\right) \frac{\boldsymbol{M} \boldsymbol{x}}{\|\boldsymbol{M} \boldsymbol{x}\|_2}. \label{mobius matrix multiplication}
\end{align}

\noindent  \textbf{Exponential and Logarithmic Maps.} For any point $\boldsymbol{x} \in \mathcal{M}$, the bidirectional mapping between its living manifold $\mathcal{M}_{\kappa}^{n}$ and corresponding tangent space  $\mathcal{T}_{\boldsymbol{x}}\mathcal{M}_{\kappa}^{n}$ is established by the exponential map $\exp_{\boldsymbol{x}}^{\kappa}: \mathcal{T}_{\boldsymbol{x}}\mathcal{M}_{\kappa}^{n} \rightarrow \mathcal{M}_{\kappa}^{n}$ and logarithmic map $\log_{\boldsymbol{x}}^{\kappa}: \mathcal{M}_{\kappa}^{n} \rightarrow \mathcal{T}_{\boldsymbol{x}}\mathcal{M}_{\kappa}^{n}$. The clean closed-form expression is given as follows:
\vspace{-0.1in}
\begin{align}
    \exp _{\boldsymbol{x}}^{\kappa}(\boldsymbol{v}) =\boldsymbol{x} \oplus_{\kappa}&\left(\tan _{\kappa}\left(\sqrt{|\kappa|} \frac{\lambda_{\boldsymbol{x}}^{\kappa}\|\boldsymbol{v}\|_{2}}{2}\right) \frac{\boldsymbol{v}}{\|\boldsymbol{v}\|_{2}}\right), \label{exp map} \\
    \log _{\boldsymbol{x}}^{\kappa}(\boldsymbol{y}) =\frac{2}{\lambda_{\boldsymbol{x}}^{\kappa}\sqrt{|\kappa|}} &\tan _{\kappa}^{-1}\left\|-\boldsymbol{x} \oplus_{\kappa} \boldsymbol{y}\right\|_{2} \frac{-\boldsymbol{x} \oplus_{\kappa} \boldsymbol{y}}{\left\|-\boldsymbol{x} \oplus_{k} \boldsymbol{y}\right\|_{2}}. \label{log map}
\end{align}

\noindent  \textbf{Distance Metric.} In the $\kappa-$stereographic model, the distance $d_{\mathcal{M}}^{\kappa}\left(\cdot, \cdot\right)$ between two points $\boldsymbol{x}, \boldsymbol{y} \in \mathcal{M}_{\kappa}^{n}, \boldsymbol{x} \neq \boldsymbol{y}$ is defined as:
\vspace{-0.1in}
\begin{equation} \label{distance metric}
    \begin{aligned}
        d_{\mathcal{M}}^{\kappa}\left(\boldsymbol{x}, \boldsymbol{y}\right) = \frac{2}{\sqrt{|\kappa|}}\tan_{\kappa}^{-1}\left(\sqrt{|\kappa|} \left\|-\boldsymbol{x} \oplus_{k} \boldsymbol{y}\right\|_{2}\right).
    \end{aligned}
    \vspace{-0.1in}
\end{equation}
In the gyrovector formalism, $\tan_{\kappa}(\cdot)=\tanh(\cdot)$ for $\kappa <0$, otherwise, $\tan_{\kappa}(\cdot)=\tan(\cdot)$.

\subsection{Sectional Curvature and Ricci Curvature}
In Riemannian geometry, the notion of curvature describes the extent how a curve deviates from being a straight line, or a surface deviates from being a plane.
Specifically, 
sectional curvature and Ricci curvature are introduced to  describe the global and local structure, respectively \citep{Ye2020Curvature}.

\noindent \textbf{Sectional Curvature.} For each point on the manifold, sectional curvature is defined by tracing over all two-dimensional subspaces passing through the point. It is a cleaner description compared to the Riemann curvature tensor \citep{lee2018introduction}. 
Recent works \citep{ZhangWSLS21,DBLP:conf/cvpr/DaiWGJ21,DBLP:conf/acl/ChenHLZLLSZ22} usually consider the case that sectional curvature is equal everywhere on the manifold, and thus the sectional curvatures are degraded as a single constant.

\noindent \textbf{Ricci Curvature.} Ricci curvature is defined by averaging  sectional curvatures at a point.
In the literature, there are several discrete variants of Ricci curvature defined for the graphs, e.g., Ollivier-Ricci curvature \citep{ollivier2009ricci} and Forman-Ricci curvature \citep{forman2003bochner}.
The intuition of the Ricci curvature on graphs is to measure how the local geometry of an edge in the graph differs from a gird graph. 
Specifically, Ollivier version is a coarse approximation of Ricci curvature, while Forman version is combinatorial and faster to compute.

\subsection{Problem Formulation}

In this paper, we formally define a sequential interaction network as follows:


\newtheorem*{def1}{Definition (Sequential Interaction Network)} 
\begin{def1}
A sequential interaction network (SIN) is formulated a tuple $\mathcal{G} = \{\mathcal{U, I, E, T, X}\}$. $\mathcal{U}$ and $\mathcal{I}$ denote the user set and item set, respectively. Each user (item) is attached with a feature $\boldsymbol{u} \in \mathbb R^{d_\mathcal U}$ ($\boldsymbol{i} \in \mathbb R^{d_\mathcal I}$).
$\mathcal{E} \subseteq \mathcal{U} \times \mathcal{I} \times \mathcal{T}$ is the interaction set.
An interaction $e \in \mathcal{E}$ is defined  as a triplet $e = (u, i, t)$, recording that user $u \in \mathcal{U}$ and item $i \in \mathcal{I}$ interact with each other at time point $t \in \mathcal{T}$.
$\mathcal{X} \in \mathbb{R}^{|\mathcal{E}| \times d_{\mathcal{X}}}$ is the matrix summarizing the attributes of  each interaction, where $d_{\mathcal{X}}$ is the dimensionality of the attributes. 
\end{def1}

Now, we define the problem of  \emph{Self-supervised Representation Learning for Sequential Interaction Networks} as follows:
\newtheorem*{def2}{Problem Definition} 
\begin{def2}
For a sequential interaction network $\mathcal{G}$, 
it aims to learn encoding functions, 
 mapping users/items to the representation space,
   which is able to model the difference between users and item \emph{(the first issue)} and the evlovement over time \emph{(the second issue)}.
Formally, we have $\Phi_u: \mathcal{U} \rightarrow \mathbb{U}_{\kappa_u}$ and $\Phi_i: \mathcal{I} \rightarrow \mathbb{I}_{\kappa_i}$, 
where  $\mathbb{U}_{\kappa_u}$ and $\mathbb{I}_{\kappa_i}$ are two different Riemannian spaces modeling the structural patterns underlying the users and items, respectively. The curvatures $\kappa_u$ and $\kappa_i$ are the functions of time modeling the space evolvement over time.
No label information is required \emph{(the first issue)} to learn the encoding functions $\Phi_u$ and  $\Phi_i$ so as to predict future interactions with user/item embeddings.
\end{def2}
In short, we are interested in self-supervisedly representing the sequential interaction network on a couple of Riemannian user space and Riemannian item space, so that future interaction can be predicted with the user/item embeddings.

\begin{figure}[t]
    \centering
    \includegraphics[width=1.05\linewidth]{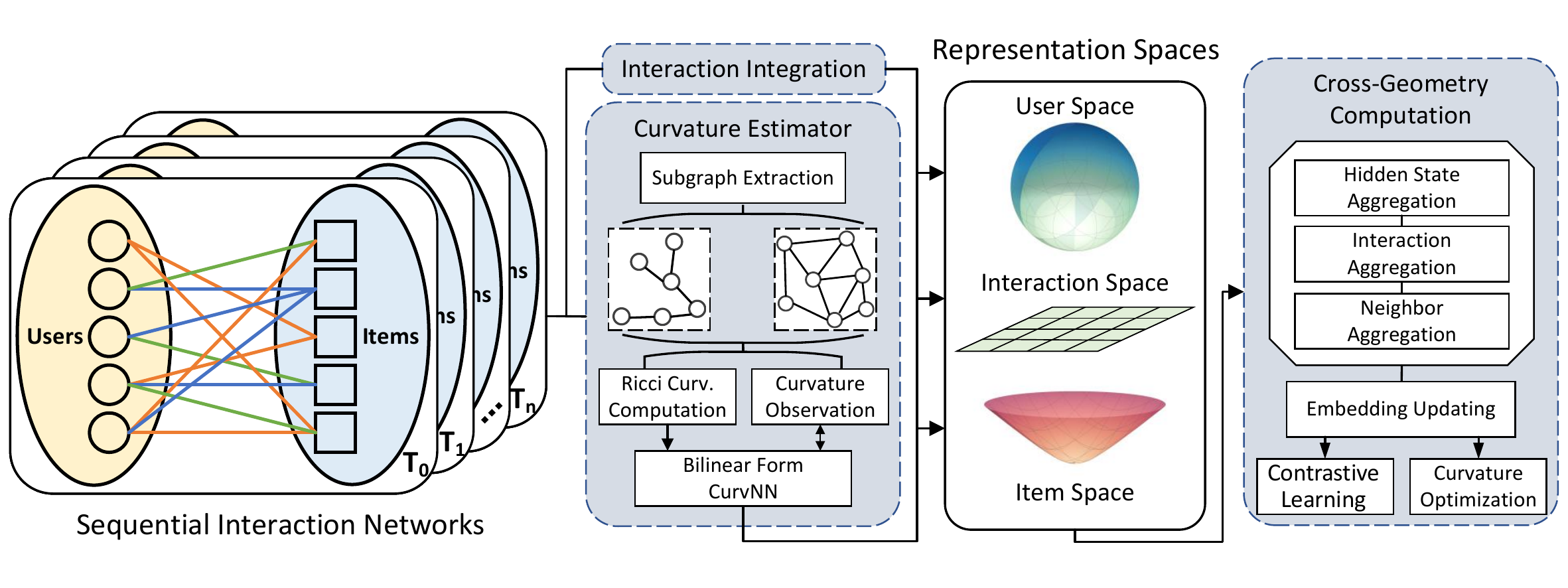}
    \caption{Illustration of \textbf{Co-Evolving GNN}. In practice, a sequence of interactions is divided into several batches according to the associated timestamps (denoted by different colors), and each batch is regarded as a time interval. Cross-Space Aggregation, Interaction Integration and the neural Curvature Estimator  are elaborated in Sec. 3.1, Sec. 3.2 and Sec. 3.3, respectively. Accordingly, user/item embeddings and curvature are learned via the proposed contrastive learning approach and curvature optimization, respectively.}
    \label{architecture}
\end{figure}

\section{Co-Evolving Graph Neural Network}

As shown in the examples and discussion in Sec. of Introduction,  users and items present inherent difference, and the pattern evolves over time.
Rather than a single and fixed-curvature representation space, we for the first time introduce the a couple of Riemannian manifolds whose curvature co-evolves over time, addressing the first and second issues.
The novel representation space is referred to as \textbf{co-evolving Riemannian manifolds}, which is one of the core contribution of our work.
Specifically, users and items are represented in the respective Riemannian manifolds, and the two manifolds are linked by a Euclidean tangent space, representing the interactions. The space evolvement over time is guided by a neural curvature estimator.
The overall framework of Co-evolving GNN is presented in \cref{architecture}. 

 Before detailing Co-evolving GNN, we collect the main notations in \cref{tab:symbols}. For the sake of clarity, we omit the  subscript when no ambiguity will occur. 

\begin{table}[h] 
    \vspace{-0.05in}
    \caption{Glossary of Main Notations in \ourmethod} \label{tab:symbols}
    \centering
    \begin{tabular}{c | l}
         \toprule
         Notation & Description  \\
         \midrule
         $\mathbb{U}_{\kappa_u}$ & Riemannian user space associated with functional curvature $\kappa_{u}$ w.r.t. time \\  
          $\boldsymbol{u}_j(T)  \in \mathbb R^{d_U}$ & $d_U-$dimensional user embedding of user $j$ during $T$ \\ 
         $\mathbb{I}_{\kappa_i}$ & Riemannian item space associated with functional curvature $\kappa_{i}$ w.r.t. time  \\  
          $\boldsymbol{i}_k(T)  \in \mathbb R^{d_U} $ & $d_I-$dimensional item embedding of item $k$ during $T$ \\
         $T_n$ & Time interval from $t_{n-1}$ to $t_n$   \\
         $\mathcal{N}_{u/i}^{T}$ & The set of neighbors centered at user $u$ (item  $i$)  during $T$ \\
         $\kappa_{u/i}^{T}$ & The value of (sectional) curvature of Riemannian user/item space during $T$    \\
           $\mathcal{E}_{u/i}^{T}$ & The set of interactions linked to user $u$ (item  $i$) during $T$ \\
         \bottomrule
    \end{tabular}
                \vspace{-0.1in}
\end{table}

              \vspace{-0.1in}
\subsection{Cross-Space Aggregation for User and Item Modeling} 

In Co-evolving GNN, we propose to model the users and items in two different representation space: user space and item space. 
They are different Riemannian spaces of $\kappa$-stereographic model. 
User and item embeddings are collected in the matrices $\boldsymbol{U}$ and $\boldsymbol{I}$,  respectively.

\label{geometry cross}
\begin{figure}[t]
    \centering
    \includegraphics[width=0.66\linewidth]{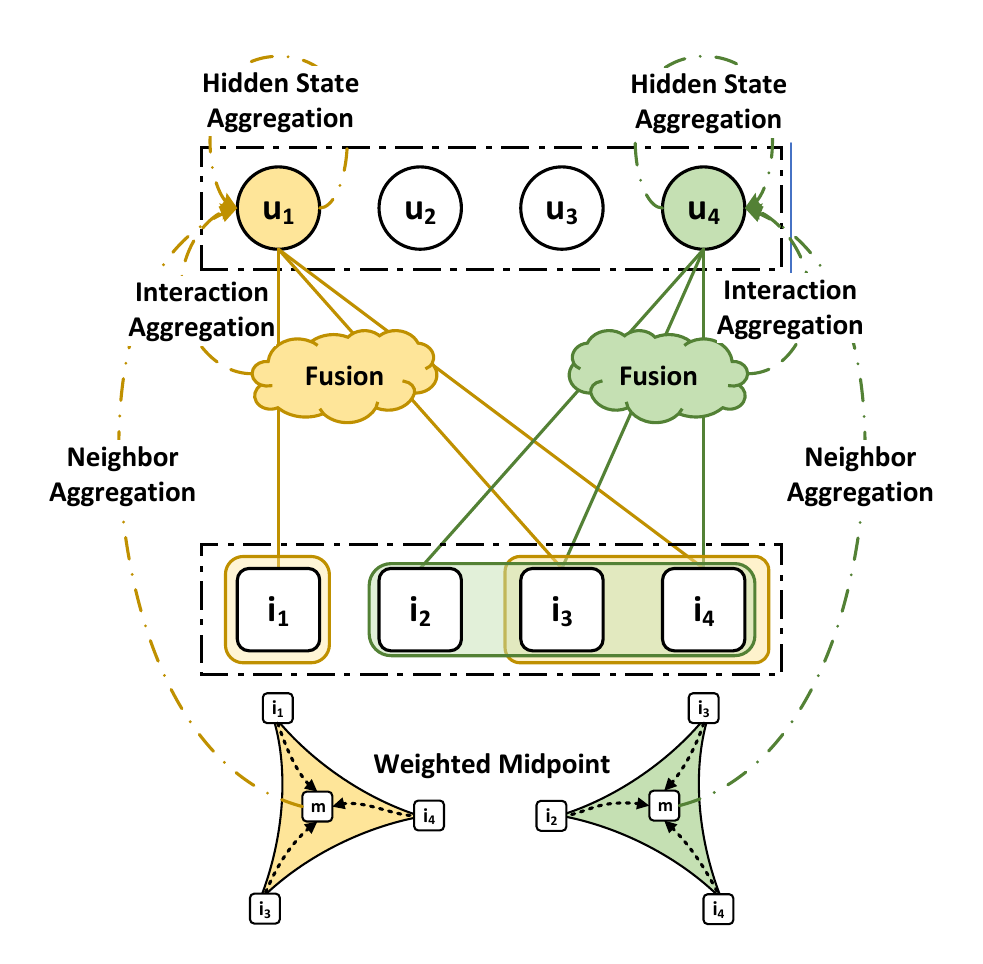}
              \vspace{-0.1in}
    \caption{Illustration of Cross-Space Aggregation. We give an example of  utilizing Eq. (\ref{user_aggregation}) to  update user embedding once.}
    \label{one step aggregation}
\end{figure}


Typical message-passing in GNNs operates on a single representation space, either the classic Euclidean or the recent hyperbolic space. 
In contrast, message-passing operates on a couple of different Riemannian spaces in our design, but how to pass the message cross different spaces still largely remains open.
To bridge this gap, we formulate the Cross-Space Aggregation with the  gyrovector formalism.
The formulation for user space and item space are given as follows,
\begin{equation}
    \boldsymbol{u}_{j}(T_n) \leftarrow \underbrace{\boldsymbol{M}_1 \otimes_{\kappa_u^T} \boldsymbol{h}_{u_j}(T_n)}_{\textit{user hidden aggregation}} 
     \oplus_{\kappa_u^T}  \underbrace{\boldsymbol{M}_2 \otimes_{\kappa_u^T} \exp_{\boldsymbol{o}}^{\kappa_u^T}(\boldsymbol{e'}_{u_j}^T)}_{\textit{interaction aggregation}}  
     \oplus_{\kappa_u^T} \underbrace{\boldsymbol{M}_3 \otimes_{\kappa_u^T} \operatorname{map}_{\kappa_i^T}^{\kappa_u^T}(\boldsymbol{i'}_{u_j}^T)}_{\textit{item aggregation}}, 
    \label{user_aggregation}
\end{equation}
\vspace{-0.1in}
\begin{equation}
    \boldsymbol{i}_{k}(T_n) \leftarrow \underbrace{\boldsymbol{M}_4 \otimes_{\kappa_i^T} \boldsymbol{h}_{i_k}(T_n)}_{\textit{item hidden aggregation}} 
     \oplus_{\kappa_i^T} \underbrace{\boldsymbol{M}_5 \otimes_{\kappa_i^T} \exp_{\boldsymbol{o}}^{\kappa_i^T}(\boldsymbol{e'}_{i_k}^T)}_{\textit{interaction aggregation}}  
     \oplus_{\kappa_i^T} \underbrace{\boldsymbol{M}_6 \otimes_{\kappa_i^T} \operatorname{map}_{\kappa_u^T}^{\kappa_i^T}(\boldsymbol{u'}_{i_k}^T)}_{\textit{user aggregation}}, \label{item_aggregation}
\end{equation}
where the matrices $\boldsymbol{M}$s are introduced for dimension transformation.
The function $\operatorname{map}_{\kappa_1}^{\kappa_2}\left(\cdot\right)$ maps the vector living in the manifold of curvature $\kappa_1$ to the manifold of curvature $\kappa_1$ with the common reference (i.e., point of north pole of the $\kappa$-stereographic model).
We provide a graphical illustration in \cref{one step aggregation} to show our idea.
Next, we detail the component of Cross-Space Aggregation of Eq. (\ref{user_aggregation}) as follows:

\begin{itemize}
\item  \textbf{The first component} updates the original  hidden representation of user $j$.
\item \textbf{The second component} aggregates the interactions associated with  user $j$. Either \textit{Early Fusion} or \textit{Late Fusion} is acceptable for interaction aggregation, i.e., $\boldsymbol{e'}_{u/i}^T = \textsc{Mlp}\left(\textsc{Pooling}\left(\boldsymbol{e} \in \mathcal{E}_{u/i}^{T}\right)\right)$ or  $\boldsymbol{e'}_{u/i}^T = \textsc{Pooling}\left(\textsc{Mlp}\left(\boldsymbol{e} \in \mathcal{E}_{u/i}^{T}\right)\right)$. The interaction modeling is introduced in the following subsection. 
\item \textbf{The third component}  aggregates items information interacted with user $j$. Here, we utilize the well-defined \textbf{weighted gyro-midpoints} \citep{ungar2010barycentric}. Concretely, the item aggregation surrounding user $u$ is formulated as
\begin{equation} \label{weighted midpoint of item}
        \boldsymbol{i'}_{u}^T = \textsc{Midpoint}_{\kappa_i^T}  \left(x_{j} \in \mathcal{N}_u^T\right) = \frac{1}{2} \otimes_{\kappa_i^T} 
         \left(\sum_{x_{j} \in \mathcal{N}_u^{T}} \frac{\alpha_{k} \lambda_{\boldsymbol{x}_{j}}^{\kappa}}{\sum_{x_{k} \in \mathcal{N}_u^{T}} \alpha_{k}\left(\lambda_{\boldsymbol{x}_{k}}^{\kappa}-1\right)} \boldsymbol{x}_{j}\right),
\end{equation}
where $\lambda_{\boldsymbol{x}}^{\kappa}$ is conformal factor introduced in Sec. 2.2. $\alpha_k$ is the weighting factor, and thus Eq. (\ref{weighted midpoint of item}) is ready to incorporate with any off-the-shelf attention mechanism in $\kappa$-stereographic model.
\end{itemize}
\textit{Note that, the explicit interaction between the nodes of different types is captured in the Cross-Space Aggregation itself while the implicit interaction between the nodes of the same type is captured by stacking multiple layers, alleviating the issue of implicit interaction ignorance.}


After the cross-space aggregation, we map user/item embeddings to next-period manifolds, and the updating rule is formulated as  as follows:
\begin{equation} \label{user updating}
    \boldsymbol{u}_j(T_{n+1}) = \operatorname{map}_{\kappa_u^{T_n}}^{\kappa_u^{T_{n+1}}} \left(\boldsymbol{u}_j\left(T_n\right)\right), 
\end{equation}
\begin{equation} \label{item updating}
    \boldsymbol{i}_j(T_{n+1}) = \operatorname{map}_{\kappa_i^{T_n}}^{\kappa_i^{T_{n+1}}} \left(\boldsymbol{i}_j\left(T_n\right)\right),
\end{equation}
where  $\kappa^{T_{n+1}}$  is the next interval curvature obtained via \cref{CurvNN}.

\subsection{Temporal Interaction Integration}

Users and items are placed in different spaces, but they are presented as a whole with the temporal interactions.
Thus, we utilize the common tangent space of the two Riemannian manifolds to model the interactions.
In other words, user space and item space are linked by a Euclidean space, the the common tangent space.

Specifically, we first perform time encoding to encode the temporal information in the timestamps.
 In other words, we are interested in a function of time encoder 
 $\Phi_t: \mathcal{T} \rightarrow \mathbb{R}^d$ that encodes a time point $t_k \in \mathcal{T}$ to a $d$-dimensional  vector $\phi(t) \in \mathcal{T}$ in Euclidean space. 
We employ the well-defined \textit{harmonic encoder} \citep{Xu2020Inductive} formulated as follows: 
\begin{equation} \label{time encoder}
    \begin{aligned}
        \phi(t) = \sqrt{\frac{1}{d}}\left[\cos\left(\omega_1 t + \theta_1\right), \cos\left(\omega_2 t + \theta_2 \right), \cdots , \cos\left(\omega_d t + \theta_d\right)\right], 
    \end{aligned}
\end{equation}
where $\omega$ and $\theta$ are the  \textbf{learnable parameters} to construct the time encoding.
Then, for each interaction $e_k \in \mathcal{E}$,  we perform  \emph{Interaction Integration} which integrates the timestamp and attribute with the following formulation,
\begin{equation} \label{edge initial embedding}
    \begin{aligned}
        \boldsymbol{e}_k = \sigma(\boldsymbol{W}_7 \cdot \left[\boldsymbol{X}_k \ : \ \phi(t_k)\right]),
    \end{aligned}
\end{equation}
where  $\phi(t_k)$ is the time encoding of the timestamp. $\left[\cdot : \cdot\right]$ denotes the concatenation. $\sigma(\cdot)$ denotes the nonlinearity and $\boldsymbol{W}_7$ is the parameter. 

\subsection{Curvature Estimator}

The previous works in the literature model sequential interaction networks in a \emph{fixed} curvature space, either zero or a negative constant. However, the fact is that the network as well as its underlying structural pattern \emph{evolves} over time. In Riemannian geometry, structural pattern of a graph is characterized as curvature.
That is, it is required to estimate the curvature to model the space evolvement over time.

In graph domain, the discrete curvatures such as Ricci curvature are introduced on the graph where the nodes are directly connected by the links (e.g., citation networks and social network).
However, SIN is a different case that the users or the items are not directly connected, i.e., SIN is bipartite.
We propose to bridge this gap by extracting a subgraph among the nodes of the same type.
The rule is that two users (items) are linked to each other if they are interacted with $K$ same item (user).
We further take sampling to accelerate the extraction process in practice.


We start with the definition of Ricci curvature $\kappa_{r}(x, y)$ to estimate curvature. 
Concretely, for any edge $(x, y)$, its  Ricci curvature of  Ollivier version \citep{ollivier2009ricci} is defined as
\begin{equation} \label{ollivier ricci}
    \begin{aligned}
        \kappa_{r}(x, y)= 1 - \frac{W\left(m_{x}, m_{y}\right)}{d_{\mathcal{M}}^{\kappa}(x, y)},
    \end{aligned}
\end{equation}
where $W(\cdot, \cdot)$ is Wasserstein distance between two mass distributions.
For node $x$, we have $m_{x}^{\alpha}\left(x_{i}\right)$, the mass distribution defined over its one-hop neighboring nodes $\mathcal{N}(x)=\{x_1, x_2, ..., x_l\}$.
\begin{equation} \label{mass distribution}
    \begin{aligned}
        m_{x}^{\alpha}\left(x_{i}\right)= \begin{cases}\alpha & \text { if } x_{i}=x, \\ (1-\alpha) / l & \text { if } x_{i} \in \mathcal{N}(x), \\ 0 & \text { otherwise. }\end{cases}
    \end{aligned}
\end{equation}
where we have  $\alpha=0.5$ following the previous works \citep{Ye2020Curvature, sia2019ollivier}. 
We collect Ricci curvatures of a time interval in the vector $\boldsymbol{r}$. 
According to the bilinear relationship between Ricci curvature and sectional curvature, 
we design a neural curvature estimator, referred to as \textit{CurvNN}, to estimate the global curvature as follows, 
\begin{equation} \label{CurvNN}
    \begin{aligned}
        \kappa_e = MLP\left(\boldsymbol{r}\right)^\top \boldsymbol{W}_8 MLP \left(\boldsymbol{r}\right),
    \end{aligned}
\end{equation}
where $MLP$ is short for Multi-Layer Perceptron and $\boldsymbol{W}_8$ is a parameter.
Note that, the formulation in Eq. (\ref{CurvNN}) is able to learn the curvature of any sign without loss of generality.

\IncMargin{1em}
\begin{algorithm}[ht]
    \caption{Observed Curvature}
    \label{curvature observation}
    \SetKwInOut{Input}{\textbf{Input}}
    \SetKwInOut{Output}{\textbf{Output}}
    \Input{An undirected graph $G$, iterations $n$}
    \Output{Observed curvature $\kappa_o$}
    \For{$m \in G$}{
        \For{$i = 1, ..., n$}{
           $b, c \in \mathop{Sample}(\mathcal{N}(m))$ and $a \in \mathop{Sample}(G)/\{m\}$\;
           Calculate $\gamma_{\mathcal{M}}(a, b, c)$\;
           Calculate $\gamma_{\mathcal{M}}(m;b, c;a)$\;
        }
        Let $\gamma(m) = \mathop{MEAN}(\gamma_{\mathcal{M}}(m;b, c;a))$\;
    }
    \textbf{Return:} $\kappa_o = \mathop{MEAN}(\gamma(m))$
\end{algorithm}

Meanwhile, we utilize the observation of the curvature to learn \textit{CurvNN}. 
Concretely, we leverage the Parallelogram Law in Riemannian space \citep{gu2018learning, fu2021ace, bachmann20a} to observe the sectional curvature. 
With a geodesic triangle $abc$ on the manifold, we have the equations below,
\begin{equation}
\begin{aligned}
\gamma_{\mathcal{M}}(a, b, c) = 
& d_{\mathcal{M}}(a, m)^2 +\frac{ d_{\mathcal{M}}(b, c)^2}{4} 
 - \frac{d_{\mathcal{M}}(a, b)^2 + d_{\mathcal{M}}(a, c)^2}{2},\\
\gamma_{\mathcal{M}}(m;b, c;a) =
& \frac{\gamma_{\mathcal{M}}(a, b, c)}{2d_{\mathcal{M}}(a, m)},
\end{aligned}
\end{equation}
where $m$ is the midpoint of $bc$.
Eq. (17) describes the extent how a triangle in the manifold deviates from being a normal triangle in Euclidean space. Recall that the notion of curvature describes the extent how a surface deviates from being a plane, and Eq. (17) is an intuitive analogy to the triangles.
Accordingly, the observed curvature $\kappa_o$ is figured out by \cref{curvature observation}, providing the supervision for \textit{CurvNN}. 
Then, the loss of curvature optimization is given as $\mathcal J_c=|\kappa_e-\kappa_o|^2$
so as to maximize the agreement between the estimated curvature and the observations.

\section{Riemannian Co-Contrastive Learning}

In this section, we present the co-contrastive learning for sequential interaction network on the Riemannian manifolds.
The novelty lies in that, in the co-contrast, user space and item space interact with each other for interaction prediction. In the meantime, we pay more attention to both hard negative and hard positive samples with the reweighing mechanism on Riemannian manifolds.

\subsection{Co-Contrast Strategy}

Contrastive learning acquires knowledge without external guidance (labels) via exploring the similarity from the data itself, 
and has achieved great success  in graph learning tasks \citep{DBLP:conf/www/YangCPLYX22,HassaniA20,QiuCDZYDWT20}.
Recently, \citep{DBLP:conf/cikm/0008YPY22,DBLP:conf/cikm/TianWSZX21} make effort on the contrastive learning for temporal networks. 
However, they are inherently different from the sequential interaction network, where the disjoint user and item set are linked by temporal interactions.
Thereby, typical contrast strategy leads to inferior performance. Concretely, contrasting the users (items) with themselves regardless of the other set tends to destroy the correlation of users and items in SIN as a whole.

\begin{figure}[t]
    \centering
    \includegraphics[width=0.6\linewidth]{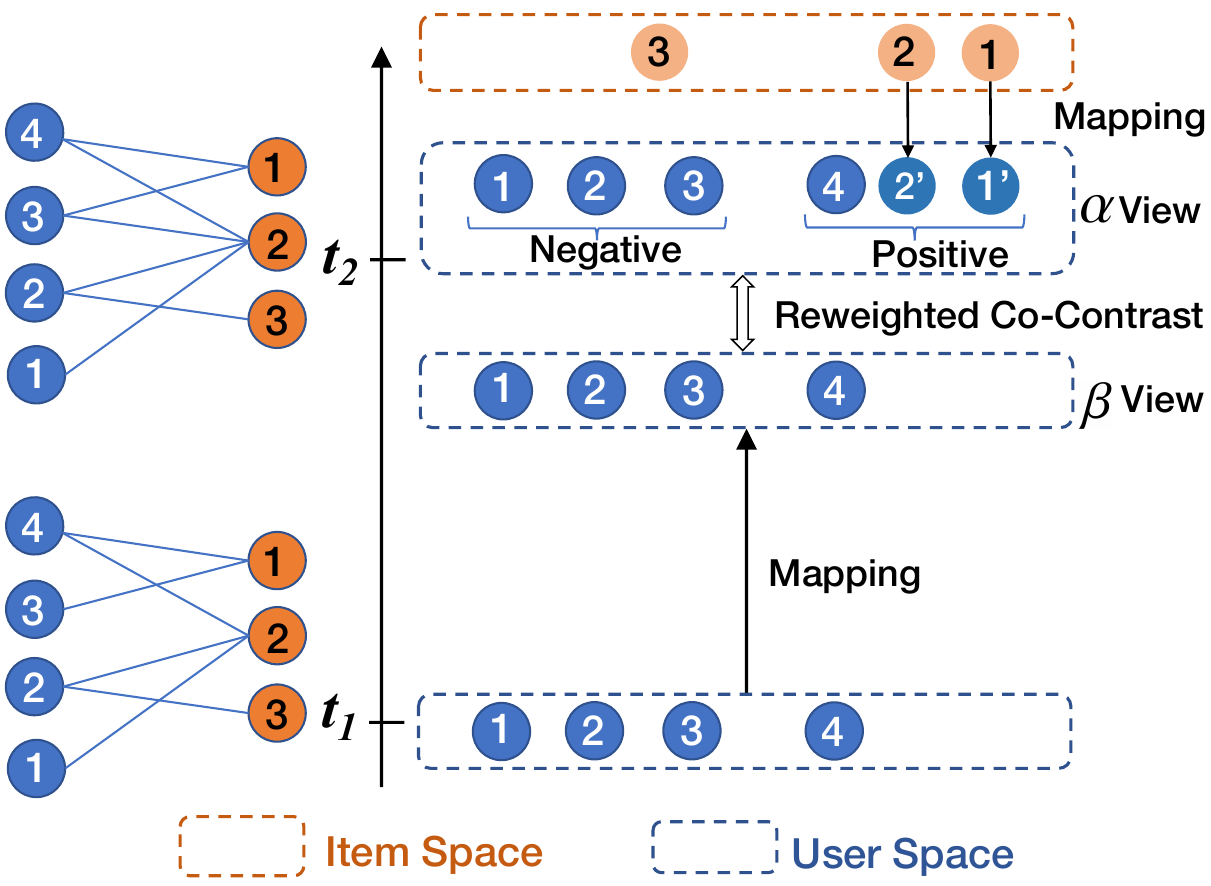}
    \caption{A toy example of the co-contrast strategy. }
    \label{toy}
\end{figure}

To bridge this gap, we introduce a co-contrast strategy, contrasting the users (items) with themselves as well as their counterparts simultaneously. 
For each anchor $x$, contrastive learning learns informative embeddings by distinguishing the positive samples ($x^+$) of $x$ from the negative ones ($x^-$) in the augmented view.
In the co-contrast strategy, \emph{we first self-augment the graph leveraging the temporal evolvement, and create different}  \textbf{temporal views}.  We take the user space for the illustration and give a toy example in Fig. \ref{toy}.  
At time $t_2$, user embeddings derived from our model term as the $\alpha$ view (denoted as $u^\alpha$). 
Those mapped from the history embedding at $t_1$ via mapping $\Gamma$ term as the $\beta$ view (denoted as $u^\beta$).
The mapping is given as 
\begin{equation}
\Gamma(\cdot)=map^{\kappa_U(t_2)}_{\kappa_U(t_1)}(\cdot,).
\label{CreateView}
\end{equation}
\emph{Second, we derive the image of the counterpart space for the co-contrast.}
In Fig. \ref{toy}, given the anchor $u_4$, we co-contrast $u_4^\alpha$ with the users and item's images in $\beta$ view. 
The negative samples are all $u^\beta_j (j \neq 4)$,
while the positive samples ($u_k^+$) are $u_4^\beta$  and the images of $i_k^\beta$ ($i_k$ are the items linking to $u_4$ at $t_2$).
The item's image is given as $map^{\kappa_U(t_2)}_{\kappa_I(t_2)}(\cdot)$.

The advantage of the proposed strategy is that, in the co-contrast, the user space interacts with the item space and vice versa, 
so as to capture user-item correlation and facilitate interaction prediction.

\subsection{Temporal Similarity on Riemannian Manifolds}

Measuring similarity in the sequential interaction network is nontrivial. 
On the one hand, Euclidean functions cannot be used on the Riemannian Manifolds. 
On the other hand, typical similarity function is time-independent \citep{abs-1807-03748,VelickovicFHLBH19,HassaniA20} but temporal information is important for sequential interaction networks.
To this end, we formulate a temporal similarity function on Riemannian manifolds as follows,
\begin{equation}
sim(x_1,x_2)=(\boldsymbol t_1^T \boldsymbol t_2) \operatorname{Sigmoid}\left(-d(\boldsymbol x_1,\boldsymbol x_2)\right),
\label{sim func}
\end{equation}
where $d(\cdot, \cdot)$ is the distance function on Riemannian manifolds, and $\boldsymbol t_i$ is the time encoding of $t_i$ derived via $\boldsymbol t_i=\phi(t_i)$ in Eq. (\ref{time encoder}).

We show that \emph{the inner product term in Eq. (\ref{sim func}) is a function of the time span} $(t_2-t_1)$, encoding the relative pattern between the samples with respect to the time (i.e., \textbf{translation invariant}). 

We start with the Bochner's Theorem as follows.
\newtheorem*{thm1}{Bochner's Theorem} 
\begin{thm1}
A translation-invariant kernel $\mathcal K(x, y) = f(x - y)$ is positive definite, where $\mathcal K$ is continuous and is defined on $\mathbb R^d$,  if and only if there exists a non-negative measure on $\mathbb R$ such that $f$ is the Fourier transformation of the measure.
\end{thm1}
Now, we prove the translation invariant  of  the inner product (Proposition 1).
\newtheorem*{pro1}{Proposition 1 (Translation Invariant)} 
\begin{pro1}
Given the time encoding in Eq. (\ref{time encoder}), there exists a real function $f$ such that the inner product of time encoding at $t_2$ and $t_1$ can be expressed as $f(t_2-t_1)$.
\end{pro1}
\begin{proof}
According to the Bochner's Theorem, 
the proposition holds if the kernel $\mathcal K$ induced by the time encoding in Eq. (\ref{time encoder}) is translation invariant. 
Given the fact of trigonometric functions, induced kernel $\mathcal K$ is given as
\begin{equation}
\begin{aligned}
\mathcal K(t_1, t_2)
=\phi(t_1)^\top\phi(t_1) 
& = \mathbb{E}_\omega\left[\cos \left(\omega t_1\right) \cos \left(\omega t_2\right)+\sin \left(\omega t_1\right) \sin \left(\omega t_2\right)\right]\\
& =\mathbb{E}_\omega\left[\cos \left(\omega\left(t_1-t_2\right)\right)\right],
\end{aligned}
\label{real part}
\end{equation}
where $\omega$ is specified in Eq. (12).
With a non-negative probability measure $p(\omega)$ on $\mathbb R$, we consider the Fourier transformation as follows,
\begin{equation}
\mathbb{E}_\omega\left[\xi_\omega\left(t_1\right) \xi_\omega\left(t_2\right)^*\right]
 =\int_{\mathbb{R}} e^{i \omega\left(t_1-t_2\right)} p(\omega) d \omega 
 =f(t_1-t_2),
\end{equation}
where $\xi_\omega(t)=e^{i \omega t}$. $i$ is the image unit, and $*$ denotes the complex conjugate. Eq. (\ref{real part}) is the real part of the Fourier transformation, and there exists a real $f$ when scaled properly \citep{Xu2020Inductive}. 
That is, the kernel $\mathcal K$ is translation invariant, and thereby $\mathcal K(t_1,t_2)=f(t_2-t_1)$.
\end{proof}
\noindent \textbf{Remark.} Note that, $f$ is learned in the contrastive learning, and additionally, the $f$ implied in Eq. (\ref{sim func}) has better expressive ability than the exponential decay, as shown in the Ablation Study in Sec. 5.



\subsection{Reweighing InfoNCE Loss}

The classic InfoNCE with a standard binary cross-entropy \citep{DBLP:conf/cikm/TianWSZX21,VelickovicFHLBH19} is given as follows, 
\begin{equation}
- \mathbb E_{x} \left[    \mathbb E_{x^+ \sim p^+}   log\frac{1}{1+e^{-sim(x,x^+)}} + 
\mathbb E_{x^- \sim p^-} log\frac{1}{1+e^{sim(x,x^-)}} \right],
\label{infoNCE}
\end{equation}
A major shortcoming of the InfoNCE loss above is that all the samples in the augmented view are treated equally, i.e., neglecting the hardness of the samples \citep{DBLP:conf/iclr/RobinsonCSJ21,DBLP:conf/icml/XiaWWCL22}. 
However, the importance of different samples tends to be different in the contrastive learning. 
More attention to the hard samples boosts the learning performances. 
A \emph{hard negative} has a similar representation to the anchor so that it is hard to be distinguished.
Similarly, in the case of positive pairs, more attention is required when the positive sample is far away from the anchor in the representation space, which is known as \emph{hard positive}.
Accordingly, we suggest the following sampling distribution for negative and positive samples, 
\begin{equation}
q^-_\eta \propto e^{\eta sim(x,x^-)}p^-, \quad
q^+_\eta \propto e^{-\eta sim(x,x^+)}p^+.
\label{sample}
\end{equation}
Note that, the nonnegative $\eta$ controls the impact of hard samples, and we have $\eta=2$ in practice.
$sim(\cdot, \cdot)$ is the similarity measure on the Riemannian manifolds.
Concretely, we introduce a normalizing constant to ensure the mass of the distribution equals to $1$, e.g., 
$q^-_\eta(x^-)=\frac{1}{Z^-} e^{\eta sim(x,x^-)}p^-(x^-)$. 

Hence, we derive the \emph{Reweighed InfoNCE Loss} with the distributions in Eq. (\ref{sample}) as follows,
\begin{equation}
- \mathbb E_{x} \left[   
\mathbb E_{x^+ \sim q_\eta ^+}   \log\frac{1}{1+e^{-sim(x,x^+)}} + 
\mathbb E_{x^- \sim q_\eta^-} \log\frac{1}{1+e^{sim(x,x^-)}}  \right].
\label{reweighInfo}
\end{equation}
Equivalently, we rewrite Eq. (\ref{reweighInfo}) with the original distributions as 
\begin{equation}
\mathbb E_{x^+ \sim q_\eta ^+}   \log\frac{1}{1+e^{-sim(x,x^+)}} =   
 \mathbb E_{x^+ \sim p^+}   \left[\frac{q_\eta ^+}{p^+} \log\frac{1}{1+e^{-sim(x,x^+)}} \right], 
\end{equation}
\begin{equation}
\mathbb E_{x^- \sim q_\eta ^-}   \log\frac{1}{1+e^{sim(x,x^-)}} =   
 \mathbb E_{x^- \sim p^-}   \left[\frac{q_\eta ^-}{p^-} \log\frac{1}{1+e^{sim(x,x^-)}} \right]. 
\end{equation}
The intuition of the proposed loss is that \emph{we up-weight the hard samples in the contrastive learning}, as shown in the right hand side of equation above.

\subsection{Reweighed Co-Contrast Loss} 
For the user space,  we formulate the reweighed co-contrast loss as follows,
\begin{equation}
\mathcal J^U_{(\alpha,\beta)}=-\frac{1}{|\mathcal U|}\sum_{i=1}^{|\mathcal U|} \left(\mathcal J^+(\boldsymbol u_i^\alpha,\boldsymbol u_k^\beta) + \mathcal J^-(\boldsymbol u_i^\alpha, \boldsymbol u_j^\beta) \right),
\label{RCCfirst}
\end{equation}
where
\begin{equation}
\mathcal J^+(\boldsymbol u_i^\alpha,\boldsymbol u_k^\beta)=\sum_{k=1}^{K^+}\frac{e^{-\eta sim(\boldsymbol u_i^\alpha, \boldsymbol u_k^\beta )}}{Z^+}\log\frac{1}{1+e^{-sim(\boldsymbol u_i^\alpha, \boldsymbol u_k^\beta)}},
\end{equation}
\begin{equation}
\mathcal J^-(\boldsymbol u_i^\alpha, \boldsymbol u_j^\beta)= \sum_{j=1}^{K^-}\frac{e^{\eta sim(\boldsymbol u_i^\alpha, \boldsymbol u_j^\beta )}}{Z^-}\log\frac{1}{1+e^{sim(\boldsymbol u_i^\alpha, \boldsymbol u_j^\beta )}}.
\end{equation}
$K^+$ and $K^-$ are the number of positive samples and negative samples, respectively.
$\{\boldsymbol u_j\}_{j \ne i}$ is the set of negative samples.
The positive sample set $\{\boldsymbol u_k\}$ consists of $\boldsymbol u_i$ and the images of items linking to $\boldsymbol u_i$. That is, \emph{we co-contrast users with themselves and the correlated items simultaneously.}
The normalizing constants are given as 
\begin{equation}
Z^+=\frac{1}{K^+}\sum_{k=1}^{K^+}e^{-\eta sim(\boldsymbol u_i^\alpha, \boldsymbol u_k^\beta )},\quad  Z^-=\frac{1}{K^-}\sum_{j=1}^{K^-}e^{\eta sim(\boldsymbol u_i^\alpha, \boldsymbol u_j^\beta )}.
\end{equation}
With the formulation above, it is obvious that the samples are \emph{reweighed by a softmax-like coefficient} so that hard samples are regulated by the learnt importance. 
Different from the previous reweighing mechanism \citep{DBLP:conf/cikm/TianWSZX21,DBLP:conf/icml/XiaWWCL22,DBLP:conf/cikm/0008YPY22}, the proposed reweighing not only regulates the hard samples in the user space iteslf, but also regulates the hard samples in the counterpart space.

Similarly, the reweighed co-contrastive loss for the items is given as follows,
\begin{equation}
\mathcal J^I_{(\alpha,\beta)}=-\frac{1}{|\mathcal I|}\sum_{i=1}^{|\mathcal I|} \left(\mathcal J^+(\boldsymbol i_i^\alpha, \boldsymbol i_k^\beta) + \mathcal J^-(\boldsymbol i_i^\alpha, \boldsymbol i_j^\beta) \right).
\label{RCClast}
\end{equation}

\begin{algorithm}[t]
    \caption{Self-supervised Learning}
    \label{training algorithm}
    \SetKwInOut{Input}{\textbf{Input}}
    \SetKwInOut{Output}{\textbf{Output}}
    \Input{A sequential interaction network $\mathcal{G} = \{\mathcal{U, I, E, T, X}\}$}
    \Output{1) User and item embeddings $\boldsymbol{U}$,$\boldsymbol{I}$ \\
                  2) A well-trained \emph{CurvNN}}
    Initialize user/item embeddings $\boldsymbol{u}_j$, $\boldsymbol{i}_k$ and map them to the respective manifold via \cref{exp map}\;
    Initialize time encoding for  interaction embeddings $\boldsymbol{e}$ via \cref{edge initial embedding}\;
    \For{$epoch = 1, ..., N$}{
        \For{ each $batch = t$}{
            Feed Ricci curvatures into \emph{CurvNN} via \cref{CurvNN}\;
            Calculate the observed curvature via Algorithm 1\;
            \textbf{Generate the $\alpha$ view} by forwarding Co-evolving GNN in Eq. (\ref{user_aggregation}-\ref{item updating})\;
            \textbf{Generate the $\beta$ view} by mapping from the history embeddings via Eq. (\ref{CreateView})\;
            Calculate the Reweighed Co-Contrast loss in Eq. (\ref{RCCfirst}-\ref{RCClast})\;
            Calculate the overall loss in Eq. (\ref{final})\;
            Back propagation, update parameters\;
            Record the curvature of batch $t$\;
            }
        }
\end{algorithm}

\noindent \textbf{Overall Loss of \ourmethod.} Finally, we formulate the overall loss of \ourmethod below,
\begin{equation}
\mathcal L=\left(\mathcal J^U_{(\alpha,\beta)}+\mathcal J^U_{(\beta,\alpha)}\right) + w_1\left(\mathcal J^I_{(\alpha,\beta)}+\mathcal J^I_{(\beta,\alpha)}\right)+w_2 \mathcal J_c,
\label{final}
\end{equation}
where the $\alpha$ view is contrasted with  the $\beta$ view, and vice versa.
 $\mathcal J_c$ is the loss of curvature optimization in Sec. 3.3, and the $w$'s are weighting coefficients.

\emph{In summary,} \ourmethod \emph{learns user and item embeddings on co-evolving Riemannian manifolds, where user space and item space interact with each other in the (Reweighed) Co-Contrastive Learning for interaction prediction.}

\subsection{Computational Complexity Analysis}
The procedure to train our \ourmethod is summarized in Algorithm 2. 
The most expensive component of our model is the Reweighed Co-Contrast, whose computational complexity is $O(|\mathcal{U}| (|\mathcal{U}| + D_U) +|\mathcal{I}| (|\mathcal{I}| + D_I))$.
Concretely, the former term is the complexity of the contrastive learning in the user space, and $D_U$ is the user's maximum degree. The degree here means the number of items linking to the user.
The latter one is the complexity of the contrastive learning in the item space, and $D_I$ is the item's maximum degree.
In the curvature estimation, the most expensive component is to solve the Wasserstein distance sub-problem implied in the Olliver-Ricci curvature. The computational complexity is $O(V^3\log V)$, where $V$ is the number of the nodes in the subgraph introduced in Sec 3.3.
It is noteworthy to mention that the Olliver-Ricci curvature can be effectively obtained following  \cite{CDRiccFlow,DBLP:conf/iclr/YeLM0020}. 
In addition, to avoid repeated calculation of Olliver-Ricci curvature, we adopt an \textbf{Offline} computation and saves it in advance as a pre-processing for each batch.

\section{Experiments} \label{Experiments}

In this section, we compare the proposed \ourmethod with $10$ strong baselines on $5$ public datasets with the aim of answering the research questions as follows (\emph{RQs}):
\begin{itemize}
  \item \textbf{\emph{RQ1}}: How does the proposed \ourmethod perform?
  \item \textbf{\emph{RQ2}}: How does the proposed component contributes to the success of \ourmethod?
    \item \textbf{\emph{RQ3}}: How is \ourmethod sensitive to the hyperparameters?
\end{itemize}


\subsection{Experiment Setups}

\subsubsection{Datasets}
 To examine the performance of \ourmethod, we conduct experiments on $\boldsymbol 5$ real-world datasets: \textbf{MOOC}, \textbf{Wikipedia}, \textbf{Reddit}, \textbf{LastFM} and \textbf{Movielen} \citep{jodie,CAW,hili}. The statistics of the datasets are detailed in Table \ref{tab:Statistics of datasets}.

 \begin{table}[h] 
    \caption{Statistics of datasets} 
    \vspace{0.05in}
    \centering
    \begin{tabular}{l|cccc}
         \toprule
         Dataset & \#Users & \#Items & \#Links & \#Features \\
         \midrule
         MOOC & 7,047 & 97 & 411,749 & 4 \\
         Wikipedia & 8,227 & 1,000 & 157,474 & 172 \\
         Reddit & 10,000 & 984 & 672,447 & 172 \\
         LastFM & 980   & 1000 & 1,293,103 & 0 \\
         Movielen & 610 & 9,725 & 100,836 & 300 \\
         \bottomrule
    \end{tabular}\label{tab:Statistics of datasets}
\end{table}


\subsubsection{Baselines} 
To evaluate the effectiveness of our model, we compare our model with $\boldsymbol{10}$ state-of-the-art baselines, which are categorized as follows: 
\begin{itemize}
    \item \textbf{Recurrent models:} We compare with a family of RNNs designed for sequence data: LSTM, T-LSTM \citep{time-lstm}, RRN \citep{RRN}.
    \item \textbf{Random walking models:} CAW \citep{CAW}, CTDNE \citep{CTDNE} are two temporal network models. The former adopts causal and anonymous random walks.
    \item \textbf{Interaction models:} JODIE \citep{jodie}, HILI \citep{hili} and DeePRed \citep{deepred} are three state-of-the-art methods employing recursive network to model the user-item interaction.
    \item \textbf{Hyperbolic models:} HGCF \citep{HGCF} is a hyperbolic method on collaborative filtering. 
\end{itemize}

\noindent Note that, none of the existing studies consider SIN learning on the generic Riemannian manifolds, to the best of knowledge. We the first time bridge this gap to learn SIN on the co-evolving Riemannian manifolds.
Also, we include \emph{the conference version} \textsc{Sincere} \citep{YeJD23WWW} as a baseline, and present the detailed comparison between the proposed model  and \textsc{Sincere} in Sec. 5.2.

\subsubsection{Evaluation Metrics} 
In this paper, we employ two metrics, Mean Reciprocal Rank ($MRR$) and $Recall@k$ to measure the performance of all methods. 
\begin{itemize}
\item  $Recall@k$ is defined as $=\frac{1}{N}\sum\nolimits_{i = 1}^N\mathbb{I}(rank_i<=k)$, where $\mathbb{I}(\cdot)$ is the indicator function. $Recall@k$ means the true items appear in the top $k$ of sorted relevant item candidates.
\item $MRR$ is defined as  $\frac{1}{N}\sum\nolimits_{i = 1}^N {\frac{1}{{rank_i}}}$, where $rank_i$ is the rank of predicted $i$-th item, and $N$ is the amount of all items.  Accordingly, $MRR$ highlights the ranking of the prediction, and performs better than mean rank due to its stability.
\end{itemize}
The higher the metrics, the better the performance. 

\subsubsection{Implementation Details} 

The baselines are implemented with settings of the best performance according to the original papers.
We conduct the $80\%-10\%-10\%$  train-valid-test split with the chronological order of the interaction samples
If user/item feature are Euclidean, we map the features to respective Riemannian space via the logarithmic map.
In \textbf{C}\textsc{Sincere}, we have $w_1=1$ to balance the contrast between user space and item space.  $w_2=10$ so as to highlight the curvature learning. 
$\eta=2$ in order to highlight the hard samples in the reweighting mechanism in our co-contrast approach.
The embedding dimension is set to $64$ as default.
Given the parameters live in the Euclidean tangent space with our design, \textbf{C}\textsc{Sincere} is optimized at ease, and the Adam optimizer is employed where the learning rate is $0.001$ while the dropout rate is $0.3$.
For the Riemannian baselines (i.e., HGCF, \textsc{Sincere} and \textbf{C}\textsc{Sincere}), we set set the embedding dimension as $64$ to ensure a fair comparison.
In practice, with the current batch of interactions, we first estimate the curvatures of the next batch via \emph{CurNN},
and then infer the probability of the items interacting with the target user.
A ranked list of top-$k$ items is given for interaction prediction.

\begin{table}[t]
    \caption{Future interaction prediction: Performance comparison in terms of mean reciprocal rank ($MRR$) and $Recall@10$. The best results are in \textbf{bold} and second best results are \underline{underlined}.}
    \vspace{0.05in}
    \label{tab:main-results-with-onehot}
    \centering
    \begin{tabular}{l|cc|cc|cc|cc|cc}
        \toprule
                     &\multicolumn{2}{c|}{\textbf{MOOC}}&\multicolumn{2}{c|}{\textbf{Wikipedia}}&\multicolumn{2}{c|}{\textbf{Reddit}}&\multicolumn{2}{c|}{\textbf{LastFM}}&\multicolumn{2}{c}{\textbf{Movielen}} \\
        Method &$MRR$&$Recall$&$MRR$&$Recall$&$MRR$&$Recall$&$MRR$&$Recall$&$MRR$&$Recall$ \\
        \midrule
        \midrule
        LSTM 
        & 0.055 & 0.109 & 0.329 & 0.455 & 0.355 & 0.551 & 0.062 & 0.119 & 0.031 & 0.060 \\
        T-LSTM 
        & 0.079 & 0.161 & 0.247 & 0.342 & 0.387 & 0.573 & 0.068 & 0.137 & 0.046 & 0.084 \\
        RRN 
        & 0.127 & 0.230 & 0.522 & 0.617 & 0.603 & 0.747 & 0.089 & 0.182 & 0.072 & 0.181\\
        CAW 
        & 0.200 & 0.427 & 0.656 & 0.862 & 0.672 & 0.794 & 0.121 & 0.272 & 0.096 & 0.243 \\
        CTDNE 
        & 0.173 & 0.368 & 0.035 & 0.056 & 0.165  & 0.257 & 0.010 & 0.011 & 0.033 & 0.051 \\
        JODIE 
        & 0.465 & 0.765 & 0.746 & 0.822 & 0.726 & 0.852 & 0.195 & 0.307 & 0.428 & 0.685 \\
        HILI 
        & 0.436 & 0.826 & 0.761 & 0.853 & 0.735 & 0.868 & 0.252 & 0.427 & 0.469 & 0.784 \\
        DeePRed 
        & 0.458 & 0.532 & \textbf{0.885} & \underline{0.889} & \underline{0.828} & 0.833 & 0.393 & 0.416 & 0.441 & 0.472 \\
        HGCF 
        & 0.284 & 0.618 & 0.123 & 0.344 & 0.239 & 0.483 & 0.040 & 0.083 & 0.108 & 0.260 \\
        \hline
        \hline
        \textsc{Sincere}
        & \underline{0.586} & \underline{0.885} & 0.793 & 0.865 & 0.825 & \underline{0.883} & \underline{0.425} & \underline{0.466} & \underline{0.511} & \underline{0.819} \\
        \ourmethod
        &  \textbf{0.630} & \textbf{0.912} & \underline{0.859} & \textbf{0.908}  &  \textbf{0.867}  & \textbf{0.931}  &   \textbf{0.478}   &  \textbf{0.526} &  \textbf{0.554}   & \textbf{0.831}  \\
        \bottomrule
    \end{tabular}
\end{table}

\subsection{RQ1: Future Interaction Prediction} \label{future interaction prediction}
The task of interaction prediction is to predict the next item that a user will interact with historical iterations 

We summarize the prediction results on all the datasets in terms of both $MRR$ and $Recall@10$ in \cref{tab:main-results-with-onehot}. 
Note that, we report the empirical results of the mean value of $5$ independent run for fair comparison.
As shown in \cref{tab:main-results-with-onehot}, our \ourmethod achieves the best results in most of the cases.
It shows the effectiveness of our idea,  introducing the co-evolving Riemannian manifolds to learn the sequential interaction network.
In the experiments, first, we find that the performance of the state-of-the-art JODIE and HILI has a relatively strong reliance on the time-independent component of the embeddings. 
We have reported such finding in the conference version \citep{YeJD23WWW}, and in this paper, we focus on the proposed contrastive model.
Second, among all methods, we find that \ourmethod, \textsc{Sincere} as well as DeePRed present high  $MRR$ and $Recall@10$ with the smallest gaps.
It shows that these models can not only make good prediction  but also distinguish the groundtruth item from negative ones with higher ranks. 

\textbf{Comparing SINCERE.} 
Here, we discuss on the methodology, computational complexity and, more importantly, the effectiveness.
Both the proposed model and the previous SINCERE consider the representation learning on co-evolving Riemannian manifolds and build with the co-evolving GNN.
The difference lies in the learning paradigm. SINCERE conducts the generative learning that reconstructs the temporal interactions in chronological order, while our model leverages the novel co-contrastive learning. 
In terms of computational complexity, SINCERE is in the order of $O(|\mathcal E|)$, where $|\mathcal E|$ is the number of interactions. Our model is in the order of $O(N^2)$, $N=max(|\mathcal U|, |\mathcal I|)$.
The complexity of our model is slightly higher than that of SINCERE as $\mathcal{E} \subseteq \mathcal{U} \times \mathcal{I} \times \mathcal{T}$, and is in the same order as typical contrastive graph model such as \citep{HassaniA20,QiuCDZYDWT20,VelickovicFHLBH19}.
In terms of effectiveness, it is noteworthy to mention that \textbf{our model consistently outperforms SINCERE}, showing the superiority of the reweighted co-contrastive learning. We further investigate our contrastive learning approach in the following section.

\begin{table}[t]
    \caption{Ablation study on Wikipedia and Movielen datasets.}
    \vspace{0.05in}
    \label{tab:ablation-results}
    \centering
    \begin{tabular}{c l|cc|cc}
        \toprule
        &                       &\multicolumn{2}{c|}{\textbf{Wikipedia}}&\multicolumn{2}{c}{\textbf{Movielen}} \\
    \multicolumn{2}{c|}{ \textbf{Variant}} &$MRR$&$Recall@10$&$MRR$&$Recall@10$ \\
        \midrule
        \midrule
        \multirow{5}{*}{\rotatebox{90}{\emph{Zero}} } 
       & \textsc{Sincere} & 0.692  &0.758  &0.374  &0.616  \\
      & $w/oKernel$     &0.717  &0.806  &0.442  &0.718  \\
      &  $w/oReweigh$ &0.743  &0.822  &0.439  &0.692  \\
     & $w/oCoCon$      &0.671  &0.747  &0.352  &0.605  \\
       &  \ourmethod  & \textbf{0.786} & \textbf{0.835} & \textbf{0.461}   & \textbf{0.733}  \\
            \midrule
        \midrule
          \multirow{5}{*}{\rotatebox{90}{\emph{Static}} } 
      & \textsc{Sincere}  & 0.747  &0.834  & 0.402 &0.664  \\
      &   $w/oKernel$    &0.782  &0.867  & 0.515 &0.789  \\
      &  $w/oReweigh$  &0.779  &0.849  & 0.483 &0.756 \\
      & $w/oCoCon$      &0.705  &0.852  & 0.410 & 0.713 \\
       &   \ourmethod  & \textbf{0.818} & \textbf{0.894} & \textbf{0.527}   & \textbf{0.802}  \\
            \midrule
        \midrule
        \multirow{5}{*}{\rotatebox{90}{\emph{Evolve}} } 
     & \textsc{Sincere}  & 0.793  & 0.865  & 0.511 &0.819  \\
      &   $w/oKernel$   & 0.812  &0.893  & 0.536 & 0.822 \\
      &  $w/oReweigh$ & 0.808  &0.871  & 0.529 &0.827  \\
       & $w/oCoCon$    & 0.762  &0.859  & 0.497 &0.810  \\
       &   \ourmethod  & \textbf{0.859} & \textbf{0.908} & \textbf{0.554}   & \textbf{0.831}  \\
        \bottomrule
    \end{tabular}
\end{table}

\subsection{RQ2: Ablation Study}
In this section, we study how the proposed component contributes to the success of \ourmethod.
To this end, we introduce 6 kinds of variants as follows:
\begin{itemize}
\item To evaluate the effectiveness of  Riemannian manifold (i.e., $\kappa$-stereographic spaces), we introduce the Euclidean variant by fixing the curvature to zero, denoted as $Zero$.
\item To evaluate the effectiveness of  curvature evolvement via $CurvNN$, we introduce the static variant denoted as $Static$. The curvature is estimated over the entire network regardless of time information.
\item To evaluate the effectiveness of  similarity based on transition invariant kernel, we introduce the variant by replacing the inner product in Eq. (\ref{sim func}) with an exponential decay, denoted as $w/oKernel$.
\item To evaluate the effectiveness of  Reweighing, we introduce the variant disabling the reweighing mechanism, denoted as $w/oReweigh$. That is, we utilize the InfoNCE loss on Riemannian manifolds.
\item To evaluate the effectiveness of  Co-Contrast, we introduce the variant that separately contrast each space with itself regardless of the other, denoted as $w/oCoCon$. Concretely, we remove the positive samples of the counterpart space.
\item We include the previous version of our model (ConfVer) as a reference in the discussion.
\end{itemize}

We report their performance on Wikipedia and Movielen datasets in \cref{tab:ablation-results}, and find that:
1) The models with evolving curvatures consistently outperform the models of zero and static curvatures. 
It shows that the proposed $CurvNN$ effectively captures the structural evolvement over time. 
Rather than Euclidean space or a static curvature space, \emph{the proposed co-evolving Riemannian manifold is well aligned with the sequential interaction network}, 
inherently explaining the superiority of our model and the inferiority of the baselines.
2) \ourmethod beats those $w/oKernel$. 
It shows the better expressive ability of the proposed kernel than that of the exponential decay.
3) \ourmethod has better results than those $w/oReweigh$. 
It suggests the necessity of paying more attention to the hard samples and the effectiveness of our reweighing.
4) \ourmethod has better results than those $w/oCoCon$. 
Also, we observe that $w/oCoCon$ variants may have inferior results to the previous SINCERE,
but \ourmethod with Co-Contrast consistently achieves better results than SINCERE. 
It shows that \emph{contrastive learning on interaction networks is nontrivial, verifying the motivation of our co-contrastive learning}.
The effectiveness of the proposed co-contrastive learning lies in that user space and item space interact with each other while exploring the similarity of the data itself.

\subsection{RQ3: Hyperparameter Sensitivity} \label{hyperparameter sensitivity analysis}

To investigate hyperparameter sensitivity of \ourmethod, we conduct several experiments with different settings of the hyperparameters (i.e., embedding dimension and sampling ratio for curvature estimation).

\begin{figure} 
\subfigure[MRR on Reddit]{
\includegraphics[width=0.4\linewidth]{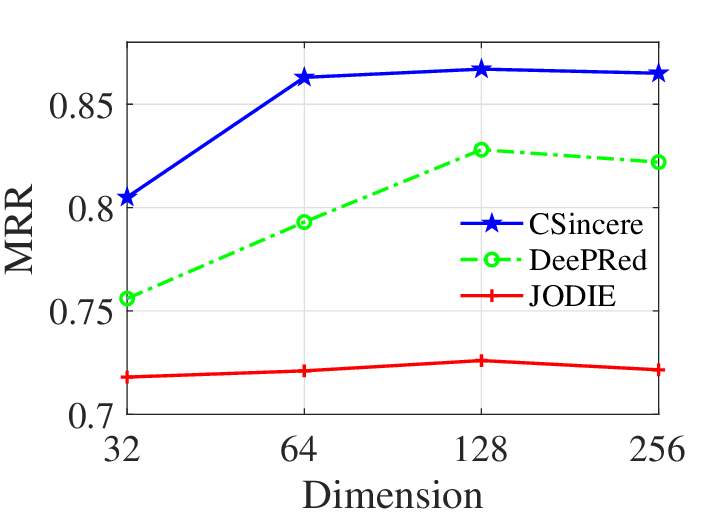}}
\subfigure[MRR on LastFM]{
\includegraphics[width=0.4\linewidth]{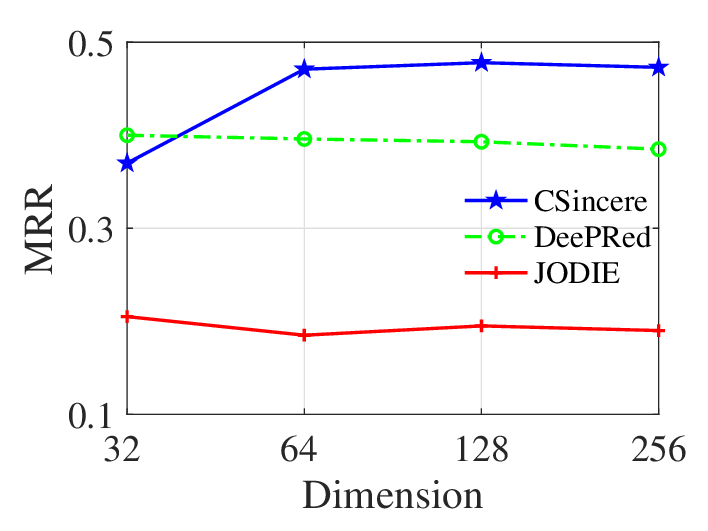}}
\vfill
    \vspace{-0.1in}
\subfigure[Recall@10 on Reddit]{
\includegraphics[width=0.4\linewidth]{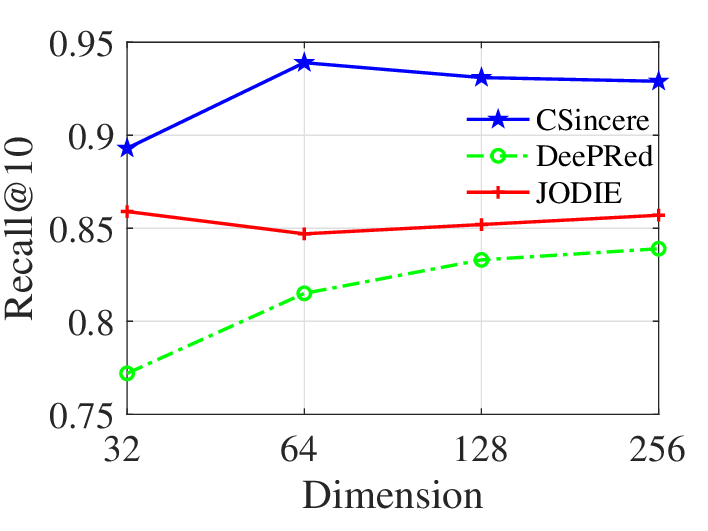}}
\subfigure[Recall@10 on LastFM]{
\includegraphics[width=0.4\linewidth]{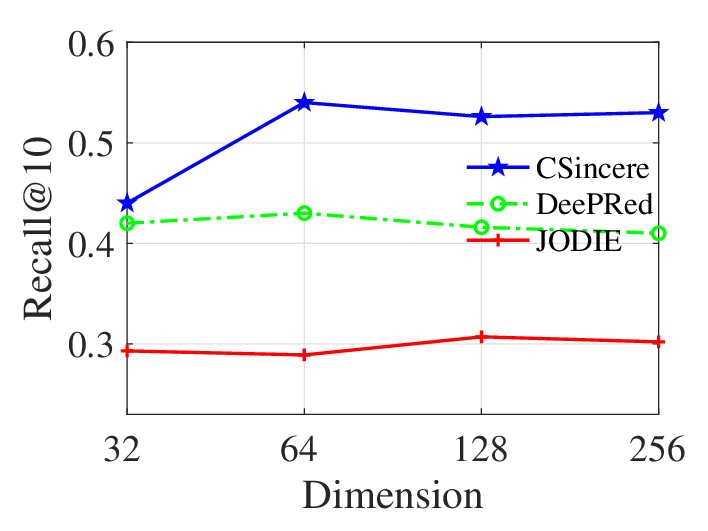}}
\caption{Effect of embedding dimensions}
    \label{fig:embedding_size}
\end{figure}

\begin{figure}
\centering
\hspace{-0.2in}
\subfigure{
    \includegraphics[width=0.3\linewidth]{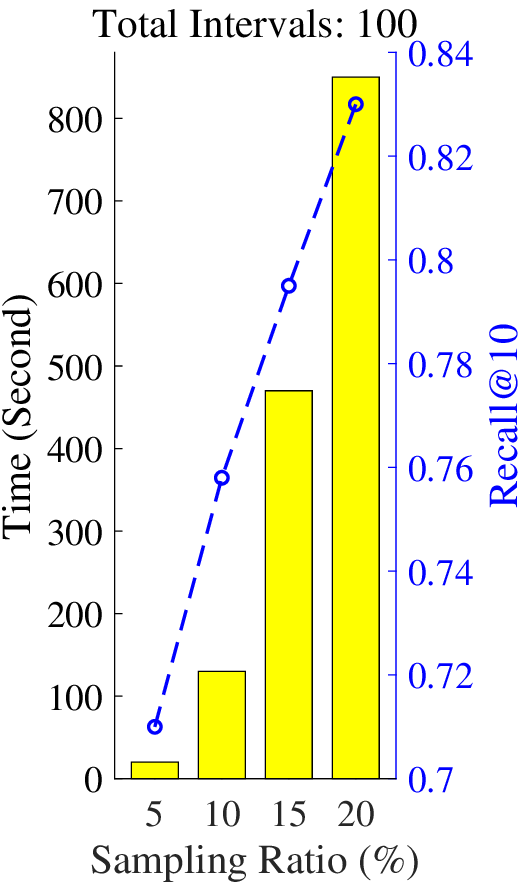}
}
\subfigure{
    \includegraphics[width=0.3\linewidth]{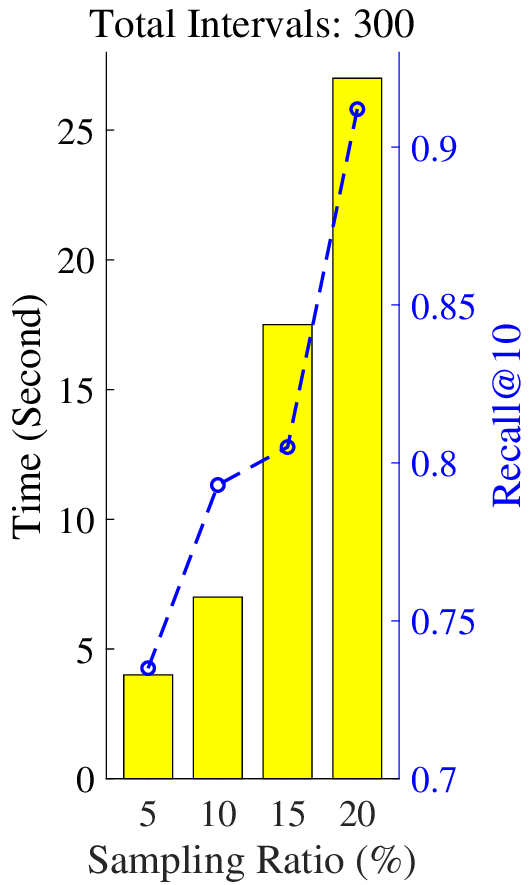}
}
\subfigure{
    \includegraphics[width=0.3\linewidth]{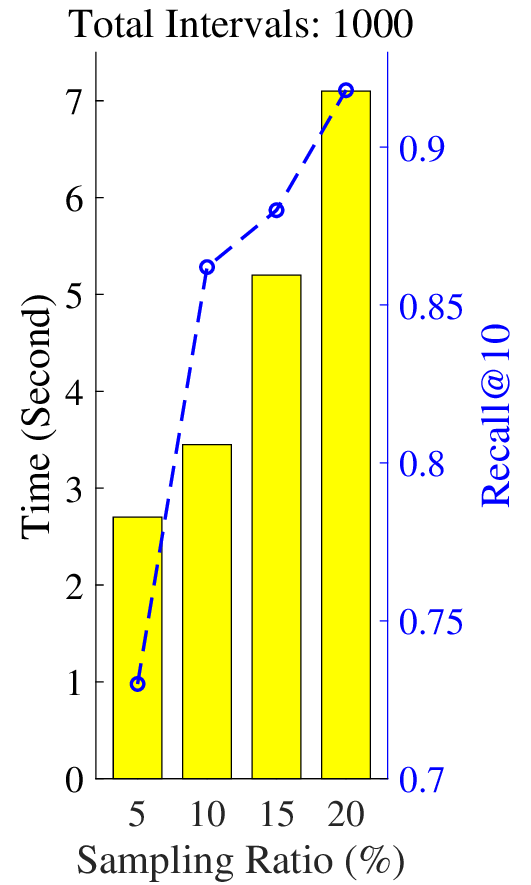}
}
\caption{Effect of sample ratio on MOOC dataset.}
\label{fig:sample_rate}
\end{figure}

First, we study the sensitivity of embedding dimension.
To this end, the embedding dimension varies in $\{32, 64, 128, 256\}$. (We use $128-$dimensional embeddings as default.)
\cref{fig:embedding_size} shows the impact of embedding dimension on the performance of our \ourmethod, DeePRed and JODIE. 
As shown in \cref{fig:embedding_size}, $128$ seems an optimal value for DeePRed in most cases. 
JODIE is relatively insensitive to embedding dimension since it counts on the static embeddings \citep{jodie}.
In some cases such as LastFM and Reddit datasets, increasing embedding dimension of \ourmethod cannot achieve further performance gain by when the dimension is larger than $64$.
In these cases, we argue that the $64-$dimensional embedding is already able to capture the information for interaction prediction, owing to the superior expressive power of the Riemannian manifold. However, if we further take the rank information of predicted items into consideration, $128-$dimensional embeddings still obtain a few performance gain,  aligned with the intuition.
To some extent, it shows the superior expressive volume of Riemannian manifolds, which is also evidenced in \citep{hnnpp,2013manifold}.
In other words, Riemannian models usually save less computational space in practice.

Second, we study the sampling ratio and the cost of time for computing Ricci curvatures. 
We set sampling ratio to $20\%$ and use $300$ intervals as default.
We summarize the performance of \ourmethod in \cref{fig:sample_rate}, where the yellow bars give the cost of time in seconds, and dashed lines show the prediction results in terms of $Recall@10$. 
In the experiment, we find that:
1) Higher sampling ratio achieves better results but sharply increases the computing time. That is, curvature estimation is of significance to interaction prediction but it is expensive.
2) The computing time evidently decreases as the number of total intervals increase. In particular, given the sampling ratio set to $20\%$, it costs over $800$ seconds with $100$ intervals, but costs less than $7$ seconds with $1000$ intervals.
Such finding suggests that, instead of increasing sampling ratio, 
higher interval number leads to less time consuming and better prediction results. 
In other words, we find a solution to tackle with the expensive curvature estimation, using higher interval number.

\section{Related Work}

\subsection{Representation Learning on SIN}
Graph representation learning maps each node on a graph to an embedding in the representation space that encodes structure and/or attribute information. 
Representation learning on SIN considers a bipartite of nodes, and learns node embeddings with a sequence of temporal interactions.
In the literature, most previous works study SINs in Euclidean space.
Among them, \emph{recurrent models} routinely find themselves owing to effectiveness on sequential data \citep{latentcross}, e.g., Time-LSTM \citep{time-lstm}, Time-Aware LSTM \citep{tlstm} and RRN \citep{RRN} model user/item dynamics with gating mechanism for long short-term memory. 
\emph{Random walking} is another line for graph representation learning.
Concretely, CTDNE \citep{CTDNE} extends the random walk to temporal networks. CAW \citep{CAW} injects the causality to inductively represent sequential networks. 
\emph{Interaction models} consider the mutual influence between users and items \citep{deepcoevolve, deepred}. HILI \citep{hili} is the successor of Jodie \citep{jodie} and both of them achieve great success. 
We noticed that researchers explore the representation learning on SIN in hyperbolic spaces.
For instance, HyperML \citep{hyperml} learns user/item encodings with the concept of metric learning in the hyperbolic space. 
HTGN \citep{HTGN} designs a recurrent architecture on the sequence of snapshots under hyperbolic geometric. 
Also, temporal GNNs achieve great success recently \citep{Xu2020Inductive, wang2021MeTA, zuo2018embedding, gupta2022tigger}, but they are different from the bipartite setting of SINs.
Recently, \citep{DBLP:journals/tkde/XiaLL23} study SIN via a probabilistic model of point process, and \citep{DBLP:conf/www/ZhangXLSJZZ23} propose a novel restart mechanism to improve the efficiency for SIN representation learning.

Please refer to \citet{kazemi2020representation, aggarwal2014evolutionary} for a more systematic reviews.
Different from the previous studies, we propose the first self-supervised SIN learning model in generic Riemannian manifold, to the best of our knowledge.

\subsection{Riemannian Graph Learning}
 Recently, Riemannian geometry (e.g., hyperbolic and spherical manifolds) emerges as a powerful alternative of the classic Euclidean ones.
A series of Riemannian graph models have been proposed.
Specifically, on \emph{hyperbolic manifolds}, shallow models are first introduced \citep{nickel2017poincare,suzuki2019hyperbolic}. Deep models, i.e., GNNs are then designed with different formulations \citep{HGCN,HGNN,ZhangWSLS21,DBLP:conf/cvpr/DaiWGJ21,DBLP:conf/acl/ChenHLZLLSZ22}. 
On \emph{constant-curvature manifolds}, $\kappa$-GCN \citep{BachmannBG20} extend GCN to  $\kappa$-sterographical model with arbitrary curvature.
On \emph{ultrahyperbolic manifolds}, a kind of pseudo Riemannian manifold, \citet{DBLP:conf/kdd/XiongZNXP0S22,pRieGNN} present GNNs in the time-space coordinates.
On \emph{quotient manifolds}, \citet{NEURIPS2021_b91b1fac} studies the entanglement of node embedding with some curvature radius.
On \emph{product manifolds}, \citet{GuSGR19,DBLP:conf/www/WangWSWNAXYC21,skopek2020mixed-curvature,sun2022self} explore informative embeddings in the collaboration of different factor manifolds.
Additionally, \citet{DBLP:conf/aaai/CruceruBG21} study the matrix manifold of Riemannian spaces.
Another line of work consider both Riemannian manifold and the Euclidean one.
For example, \citet{zhu2020graph,DBLP:conf/www/YangCPLYX22} embed the graph into the dual space of Euclidean and hyperbolic ones simultaneously. 
Recently, \citet{HTGN} and our previous work \citep{sun21aaai} model dynamic graphs in hyperbolic manifolds.
\citet{sun23aaai,sun22cikm} study the temporal evolvement of the graph in generic Riemannian manifolds.
\citet{sun23ijcai} introduces the Riemannian geometry to graph clustering.
To the best of our knowledge, we introduce \emph{the first} co-evolving Riemannian manifolds, aligning with the characteristic of SINs.

\section{Conclusion}
In this paper, we for the first time study the sequential interaction network learning on \emph{co-evolving Riemannian manifolds}, and present a novel \ourmethod. 
Concretely, we first introduce a co-evolving GNN with two $\kappa-$stereographic space bridged by the common Euclidean  tangent space, 
in which we formulate the cross-space aggregation to conduct message propagation across user space and item space, 
and design the neural curvature estimator for the space evolvement over time. 
Thereafter, we propose the Riemannian co-contrastive learning for sequential interaction networks, which in the meanwhile interplays user space and item space for interaction prediction.
Finally, extensive experiments on 5 public datasets show \ourmethod outperforms the state-of-the-art competitors.

\section*{Acknowledgments}
The authors of this paper were supported in part by National Natural Science Foundation of China under grant 62202164,  S\&T Program of Hebei through grant 20310101D, and the Fundamental Research Funds for the Central Universities 2022MS018.
Prof. Philip S. Yu is supported in part by NSF under grant  III-2106758.
Corresponding Authors: Li Sun and Mingsheng Liu.

\bibliography{custom}

\end{document}